\documentclass[twoside]{article}
\usepackage{graphicx, caption, float, amsmath, amsthm, csquotes,mathtools,amssymb,xcolor, mathrsfs, enumitem, soul, subfigure}
\setlist{nolistsep}
\usepackage[ruled]{algorithm2e}
\usepackage[accepted]{aistats2020}
\newtheorem{theorem}{Theorem}

\newtheorem{lemma}{Lemma}

\newtheorem{remark}{Remark}
\begin{document}

\twocolumn[

\aistatstitle{Laplacian-Regularized Graph Bandits: Algorithms and Theoretical Analysis}

\aistatsauthor{Kaige Yang \And Xiaowen Dong\And Laura Toni}

\aistatsaddress{University College London \And  University of Oxford \And University College London} ]

\begin{abstract}
We consider a stochastic linear bandit problem with multiple users, where the relationship between users is captured by an underlying graph and user preferences are represented as smooth signals on the  graph. 
We introduce a novel bandit algorithm where the smoothness prior is imposed via the random-walk graph Laplacian, which leads to a single-user cumulative regret scaling as $\Tilde{\mathcal{O}}(\Psi d \sqrt{T})$ with time horizon $T$, feature dimensionality $d$, and the scalar parameter $\Psi \in (0,1)$ that depends on the graph connectivity. This is an improvement over $\Tilde{\mathcal{O}}(d \sqrt{T})$ in \textbf{LinUCB}~\cite{li2010contextual}, where user relationship is not taken into account.
In terms of network regret (sum of cumulative regret over $n$ users), the proposed algorithm leads to a scaling as $\Tilde{\mathcal{O}}(\Psi d\sqrt{nT})$, which is a significant improvement over $\Tilde{\mathcal{O}}(nd\sqrt{T})$ in the state-of-the-art algorithm \textbf{Gob.Lin} \cite{cesa2013gang}. To improve scalability, we further propose a simplified algorithm with a linear computational complexity with respect to the number of users, while maintaining the same regret. 
Finally, we present a finite-time analysis on the proposed algorithms, and demonstrate their advantage in comparison with state-of-the-art graph-based bandit algorithms on both synthetic and real-world data.
\end{abstract}

\section{Introduction}
In the classical multi-armed bandit (MAB) problem, an agent  takes sequential actions, choosing  one arm  out of the $k$ available ones  and it receives an instantaneous payoff from the chosen arm only. 
The goal of the agent is to learn an action policy that maximizes the cumulative payoff over a course of $T$ rounds~\cite{robbins1952some}. 
MAB problems formalize a trade-off between exploration and exploitation, and a particular solution is imposing the principle of optimism in face of uncertainty. Specifically, the agent assigns to each arm an index called the upper confidence bound (UCB) that with high probability is an overestimate of the unknown payoff, and selects the arm with the highest index. 

Many variants of the basic MAB problem have been intensively studied, motivated by real-world applications such as ads placement and  recommender systems. In the stochastic linear bandit \cite{auer2002using}, at each round, the agent receives a hint before taking the decision. Specifically, before choosing the arm, the agent is informed of a feature vector $\mathbf{x} \in \mathbb{R}^d$ associated with each arm, referred to as the `context'.   
The payoff associated with each arm is modeled as a noisy linear function of $\mathbf{x}$ with an unknown coefficient vector $\boldsymbol{\theta} \in \mathbb{R}^d$ perturbed by a noise term $\eta$, i.e, $y=\mathbf{x}^T\boldsymbol{\theta}+\eta$, where the agent needs to learn
$\boldsymbol{\theta}$ based on the context-payoff pair $\{\mathbf{x}, y\}$ and select an arm accordingly. This problem has been well understood in the literature and many studies have already proposed asymptotically optimal algorithms \cite{auer2002using, dani2008stochastic,  agrawal2013thompson, chapelle2011empirical, lattimore2016end}. 

The problem is less understood in the case of multiple users, as opposed to a single user, where we assume a central agent needs to select arms for multiple users in a sequential fashion.
In this paper, we are interested in the setting
where there are $n$ users sharing the same set of arm choices $\mathcal{D}$ containing $m$ arms. The agent faces a set of $n$ independent instance of bandit characterized by an unknown
$\boldsymbol{\theta}_i, i \in \{1,2,...,n\}$ specific to each user. At each round, one user out of $n$ is selected uniformly at random, the agent then selects one arm from $\mathcal{D}$ for the user and receives an instantaneous payoff associated with the selected arm and the user. The overall goal is to minimize the cumulative regret (or equivalently, maximize the cumulative payoffs),
which is defined as the sum of instantaneous regret experienced by the agent over a finite time horizon $T$.  

In this setting, naively implementing bandit algorithms on each user independently will results in a cumulative regret that scales linearly with the number of users $n$. This is clearly infeasible in case of a large number of users. In many cases, however, the users are related in some way and this can be represented by a network (or graph) that encapsulates important additional source of information, such as similarities among users in terms of their preferences (user feature vectors). Exploiting this structure can mitigate the scalability problem.   
The key setup is therefore to  construct a graph where each node represents a user  and the edges identify the affinity between users. 
In real-world applications, such a graph can be a social network of users. This idea leads to a series of work on the so-called graph-based bandit problem \cite{cesa2013gang,gentile2014online,liu2018transferable, vaswani2017horde}.

Despite the previous effort, several important limitations still remain to be addressed. First,
the graph Laplacian matrix is commonly used in graph-based algorithms, but the justification of its usage remains insufficient (and as to which version of the graph Laplacian leads to optimal policies).
As a consequence, the advantage of graph-based bandit is largely shown empirically in previous works, without rigorous theoretical analysis. Furthermore, scalability remains a serious limitation of such algorithms.
Involving user graph into bandit algorithms typically results in a computational complexity that scales quadratically with the number of users, which is clearly infeasible in case of large number of users.  In this paper, we address the above limitations with the following main contributions:
\begin{itemize}
\itemsep0em 
    \item We propose a bandit algorithm \textbf{GraphUCB} based on the random-walk graph Laplacian, and show its theoretical advantages over other graph Laplacian matrices in reducing cumulative regret. We demonstrate empirically that \textbf{GraphUCB} outperforms state-of-the-art graph-based bandit algorithms in terms of cumulative regret. 
    \item As a key ingredient of the proposed algorithm, we derive a novel UCB representing the single-user bound while embedding the graph structure, which reduces the size of the confidence set, in turn leading to lower regret;
    \item To improve scalability, we further propose a simplified algorithm \textbf{GraphUCB-Local} whose complexity scales linearly with respect to the number of users, yet still holding the same regret upper bound as \textbf{GraphUCB};
    \item Finally, we derive a finite-time analysis on both algorithms and show a lower regret upper bound than other state-of-the-art graph-based bandit algorithms. 
\end{itemize}

\section{Related work}
Graph-based bandit algorithms can be roughly categorized as: $i)$ topology-based bandits, where the graph topology itself is exploited to improve learning performance, and $ii)$ spectral bandits, 
where the user feature vectors $\boldsymbol{\theta}$ are modeled as signals defined on the underlying graph, whose characteristics are then exploited in the graph spectral domain via tools provided by graph signal processing~\cite{Shuman13} to assist learning.

In topology-based bandits, the key intuition is to achieve dimensionality reduction in the user space by exploiting the graph topology. Specifically, users can be clustered based on the graph topology and a per-cluster feature vector can be learned, substantially reducing the dimensionality of the problem as opposed to the case in which one vector is learned per user. For example,  \cite{gentile2014online} clusters users based on the connected components of the user graph, and \cite{li2016collaborative} generalizes it to consider both the user graph and item graph. 
On the other hand, \cite{yang2018graph} makes use of community detection techniques on graphs to find user clusters.
More broadly, in the spirit of dimensionality reduction, even without constructing an explicit user graph, 
the work in \cite{korda2016distributed} proposes a distributed clustering algorithm

while \cite{nguyen2014dynamic} applies \textit{k-means} clustering to the user features. Despite the differences in the proposed techniques, these studies share two common drawbacks: $1)$  the learning performance  depends on the clustering algorithm being used, which tends to be expensive for large-scale graphs; $2)$ learning a per-cluster (and not per-user) feature vector means  ignoring the subtle difference between users within the same cluster. 
In short, clustering can reduce the dimensionality of the user space, but it does not necessarily preserve key users characteristics. To  achieve both goals simultaneously, there is a need for a proper mathematical framework able to
incorporate the user relationship into learning in a more direct way.

On the spectral bandit side, the strong assumption that users can be grouped  into clusters is relaxed; users are assumed to be similar with their neighbors in the graph and such similarity is reflected by the weight of graph edges.
\cite{wu2016contextual} employs a graph Laplacian-regularized estimator, which promotes similar feature vectors for users connected in the graph. In their setting, however, each arm is selected by all users jointly. This work 
results in a network regret scaling with $\Tilde{\mathcal{O}}(dn\sqrt{T})$.
\cite{vaswani2017horde} casts the same estimator as GMRF (Gaussian Markov Random Filed)  and proposes a Thompson sampling algorithm, leading to a much simpler algorithmic implementation without a  UCB evaluation. 
However, the regret bound remains $\Tilde{\mathcal{O}}(dn\sqrt{T})$.
To the best of  our knowledge, an efficient algorithm able to address the multi-user MAB problem with a sub-linear regret bound is still missing.

A proper bound and mathematical derivation of spectral MAB are provided in 
\cite{valko2014spectral}, which represents the  payoffs of arms as smooth signals on a graph with the arms being the nodes. 
Specifically, the arm features $\mathbf{x}$ are modeled as eigenvectors of the graph Laplacian and the sparsity of such eigenvectors is exploited to reduce the dimensionality of $\mathbf{x}$,  to a so-called `effective dimension' term $\Tilde{d}$. This work shows an improved regret bound $\Tilde{\mathcal{O}}(\Tilde{d}\sqrt{T})$ where $\Tilde{d}$ is significant less than $d$ in \textbf{LinUCB} \cite{abbasi2011improved}. While interesting, the proposed solution
applies to the single-user with high-dimensional arm set. 
Whereas, in our setting, the dimensionality issue is caused by the large number of users, leading to a completely different mathematical problem.

Among these works on graph bandit, the one that is most similar to our work in terms of problem definition and proposed solution is  \cite{cesa2013gang}.  In~\cite{cesa2013gang}, the graph is exploited such that each user shares instantaneous payoff with neighbors, which is promoted by a Laplacian-regularized estimator. This implicitly imposes smoothness among the feature vectors of users, resulting in the estimate of similar feature vectors for users connected by edges with strong weights in the graph. 

In our  paper, we proposed the \textbf{GraphUCB} algorithm that  builds on and improves the work of \textbf{Gob.Lin} in a number of important ways. 
\begin{itemize}
    \item First, \textbf{Gob.Lin} employs the combinatorial Laplacian as a regularizer, whereas our algorithm \textbf{GraphUCB} makes use of the random-walk graph Laplacian. We prove theoretically that the combinatorial Laplacian results in a cumulative regret scaling with the number of users, which could be large. However, random-walk graph Laplacian overcomes this serious drawback and yields a sub-linear regret with the number of users.
    \item Second, UCB used in \textbf{Gob.Lin} results in a cumulative regret that scales with $\Tilde{\mathcal{O}}(nd\sqrt{T})$ which is worse than \textbf{LinUCB} $\Tilde{\mathcal{O}}(d\sqrt{nT})$. We propose a new UCB that leads to a cumulative regret scaling with $\Tilde{\mathcal{O}}( \Psi d\sqrt{nT})$ where $\Psi \in (0,1)$. 
    \item Finally, the computational complexity of \textbf{Gob.Lin} is quadratic with respect to the number of users. Our simplified algorithm \textbf{GraphUCB-Local} scales linearly with the number of users, and at the same time enjoys the same regret bound as \textbf{GraphUCB}. This significantly improves the scalability of the proposed graph bandit algorithm.
    \end{itemize}

\section{Setting}
We consider a linear bandit problem with $m$ arms and $n$ users.
We denote  by $\mathcal{U}$  the user set with cardinality $|\mathcal{U}|=n$ and  by $\mathcal{D}$ the arm set with $|\mathcal{D}|=m$. Each arm is described by a feature vector $\mathbf{x} \in \mathbb{R}^d$, while each user is described by a parameter vector $\boldsymbol{\theta} \in \mathbb{R}^d$, with $d$ being the dimension of both vectors. The affinity between users is encoded by an undirected and weighted graph $\mathcal{G}=(V,E)$, where $V=\{1,2,.., n\}$ represents the node set for $n$ users and $E$ represents the edge set.
The graph $\mathcal{G}$ is known a priori and identified by its adjacency matrix $\mathbf{W} \in \mathbb{R}^{n \times n}$, where $W_{ij}=W_{ji}$ 
captures the affinity between
$\boldsymbol{\theta}_i$ and $\boldsymbol{\theta}_j$.
The combinatorial Laplacian of $\mathcal{G}$ is defined as $\mathbf{L}=\mathbf{D}-\mathbf{W}$, where $\mathbf{D}$ is a diagonal matrix with $D_{ii}=\sum_{i=1}^n W_{ii}$. The symmetric normalized Laplacian is defined as $\Tilde{\mathbf{L}}=\mathbf{D}^{-1/2}\mathbf{L}\mathbf{D}^{-1/2}$. In addition, the random-walk graph Laplacian is defined as $\boldsymbol{\mathcal{L}}=\mathbf{D}^{-1}\mathbf{L}$.

In our setting, the unknown user features $\boldsymbol{\Theta}=[\boldsymbol{\theta}_1,\boldsymbol{\theta}_2,...,\boldsymbol{\theta}_n]^T \in \mathbb{R}^{n \times d}$ are assumed to be smooth over  $\mathcal{G}$. 
The smoothness of $\boldsymbol{\Theta}$ over graph $\mathcal{G}$ can then be quantified using the Laplacian quadratic form of any of the three Laplacian defined above. In this work, we choose the random-walk graph Laplacian $\boldsymbol{\mathcal{L}}$ because of its two unique properties $\mathcal{L}_{ii}=1$ and $\sum_{j\neq i}{\mathcal{L}_{ij}}=-1$. The benefit of these properties
will be clear after the introduce of our proposed bandit algorithm (see Remark~\ref{remark: motivation_of_laplacian}). Mathematically, the Laplacian quadratic form based on $\boldsymbol{\mathcal{L}}$ is (see Appendix H for the derivation)
\begin{equation}
\label{eq: smoothness}
    tr(\boldsymbol{\Theta}^T \boldsymbol{\mathcal{L}} \boldsymbol{\Theta}) = \frac{1}{4} \sum_{k=1}^d \sum_{i \sim j}\left(\frac{W_{ij}}{D_{ii}}+\frac{W_{ji}}{D_{jj}}\right) \big(\Theta_{ik}- \Theta_{jk} \big)^2
\end{equation}
where $\Theta_{ik}$ is the $(i,k)$-th element of $\boldsymbol{\Theta}$. The more   the  graph $\mathcal{G}$ reflects the similarity between users correctly, the smaller the quadratic term 
$tr(\boldsymbol{\Theta}^T\boldsymbol{\mathcal{L}}\boldsymbol{\Theta})$. Specifically, $tr(\boldsymbol{\Theta}^T\boldsymbol{\mathcal{L}}\boldsymbol{\Theta})$ is small when $\Theta_{ik}$ and $\Theta_{jk}$ are similar given a large weight $\frac{W_{ij}}{D_{ii}}+\frac{W_{ji}}{D_{jj}}$.

Equipped with the above notation, we now introduce the multi-user bandit problem, in which an agent needs to take sequential decisions (e.g., recommendations) for a multitude of users appearing over time. At each time $t=1,\ldots, T$, the agent is informed about the user $i_t$ to serve, with the user being  selected uniformly at random from the user set $\mathcal{U}$. 
Then, the agent selects an arm $\mathbf{x}_t\in \mathcal{D}$ to be recommended to  user $i_t$. Upon this selection, the agent observes a payoff $y_t$, which is assumed to be generated by noisy versions of linear functions of the users and item vectors. Namely, 
\begin{equation}
    y_t=\mathbf{x}_t^T\boldsymbol{\theta}_{i_t}+\eta_t
\end{equation} 
where the noise $\eta_t$ is assumed to be $\sigma$-sub-Gaussian for any $t$.

The agent is informed about the graph $\mathcal{G}$ and the arm feature vectors $\mathbf{x}_{a}, \ a \in \{1,2,...,m\}$, while $\boldsymbol{\Theta}$ is unknown and needs to be inferred.
The goal of the agent is to learn a selection strategy that minimizes the cumulative regret with respect to an optimal strategy, which always selects the optimal arm for each user. Formally, after a time horizon $T$, the cumulative (pseudo) regret is defined as:
\begin{equation}
    R_T=\sum_{t=1}^T \bigg( (\mathbf{x}^*_{t})^T\boldsymbol{\theta}_{i_t}-\mathbf{x}_{t}^T\boldsymbol{\theta}_{i_t} \bigg)
\end{equation}
where $\mathbf{x}_t$ and $\mathbf{x}^*_{t}$ are the arm selected by the agent and the optimal strategy at $t$, respectively. Note that the optimal choice depends on $t$ as well as on the user $i_t$.   
For notation convenience in the rest of the paper, at each time step $t$, we use $i$ to generally refer to the user appeared and $\mathbf{x}_t$ to represent the feature vector of the arm selected.

\section{Laplacian-Regularized Estimator}
To estimate the user parameter $\mathbf{\Theta}$ at time $t$, we make use of the Laplacian-regularized estimator:
\begin{equation}
\label{eq: objective_function}
    \hat{\mathbf{\Theta}}_t=\arg\min_{\mathbf{\Theta}\in \mathbb{R}^{n \times d}}\sum_{i=1}^n\sum_{\tau\in \mathcal{T}_{i,t}}(\mathbf{x}_\tau^T\boldsymbol{\theta}_i-y_{\tau})^2+\alpha \ tr(\mathbf{\Theta}^T\boldsymbol{\mathcal{L}}\mathbf{\Theta})
\end{equation}
where $\boldsymbol{\theta}_i$ is the $i$-th row of $\mathbf{\Theta}$, $\mathcal{T}_{i,t}$ is the set of time steps at which user $i$ is served up to time $t$. $\mathbf{x}_{\tau}$ is the feature of arm selected by the learner, $y_{\tau}$ is the payoff from user $i$ at time $\tau$, and $\alpha$ is the regularization parameter. Clearly, Eq.~\ref{eq: objective_function} is convex and can be solved via convex optimization techniques. Specifically, it has a closed form solution \cite{alvarez2012kernels}: 
\begin{equation}
\label{eq: objective_function_vectorize}
    vec(\hat{\mathbf{\Theta}}_t)=(\boldsymbol{\Phi}_t\boldsymbol{\Phi}_t^T+\alpha \boldsymbol{\mathcal{L}}\otimes \mathbf{I})^{-1}\boldsymbol{\Phi}_t\mathbf{Y}_t
\end{equation}
where $\otimes$ is the Kronecker product, $vec(\hat{\mathbf{\Theta}}_t)\in \mathbb{R}^{nd}$ is the concatenation of the columns of $\hat{\mathbf{\Theta}}_t$,  $\mathbf{I} \in \mathbb{R}^{d \times d}$ is the identity matrix, and $\mathbf{Y}_t=[y_1, y_2,..., y_t]^T \in \mathbb{R}^t$ is the collection of all payoffs. Finally, $\boldsymbol{\Phi}_t=[\boldsymbol{ \phi}_1, \boldsymbol{ \phi}_2,..., \boldsymbol{ \phi}_t] \in \mathbb{R}^{nd \times t}$, where $\boldsymbol{\phi}_t \in \mathbb{R}^{nd}$, is a long sparse vector indicating that the arm with feature $\mathbf{x}_t$ is selected for user $i$. Formally,
\begin{equation}
\label{eq: phi}
    \boldsymbol{ \phi}_t^T=(\underbrace{0,...,0}_{(i-1)\times d \ times}, \mathbf{x}_t^T,\underbrace{0,...0}_{(n-i)\times d \ times})\,.
\end{equation}

While  Eq.~\ref{eq: objective_function} provides the estimate of feature vectors of all users at $t$, i.e., $\hat{\mathbf{\Theta}}_t$, the agent is interested in the estimation of each single-user feature vector $\hat{\boldsymbol{\theta}}_{i,t}$.  Mathematically, $\hat{\boldsymbol{\theta}}_{i,t}$ can be obtained by decoupling users in Eq.~\ref{eq: objective_function_vectorize}. This however is highly complex  due to the inversion $(\boldsymbol{\Phi}_t\boldsymbol{\Phi}_t^T+\alpha \boldsymbol{\mathcal{L}}\otimes \mathbf{I})^{-1}$, which the agent needs to preform at each time step (we recall that the Laplacian is high-dimensional). 
We notice that the close-form solution of $\hat{\boldsymbol{\theta}}_{i,t}$ can be closely approximated via a Taylor expansion of $(\boldsymbol{\Phi}_t\boldsymbol{\Phi}_t^T+\alpha \boldsymbol{\mathcal{L}}\otimes \mathbf{I})^{-1}$, as stated in   
Lemma~\ref{lemma: single_solution} and further commented and tested empirically in Appendix A.
\begin{lemma}
\label{lemma: single_solution}
$\hat{\boldsymbol{\Theta}}_t$ is obtained from Eq.~\ref{eq: objective_function_vectorize}, let $\hat{\boldsymbol{\theta}}_{i,t}$ be the $i$-th row of $\hat{\mathbf{\Theta}}_t$ which is the estimate of $\boldsymbol{\theta}_i$. $\hat{\boldsymbol{\theta}}_{i,t}$ can be approximated by\footnote{ $\mathbf{A}_{i,t}$(and $\mathbf{A}_{j,t}$) is not full-rank when $|\mathcal{T}_{i,t}| < d$. To guarantee inversion, in practice we set $\mathbf{A}_{i,t}=\sum_{\tau \in \mathcal{T}_{i,t}}\boldsymbol{x}_\tau \boldsymbol{x}^T_\tau+\lambda \mathbf{I}_d$ with $\lambda=0.01$.}:
\begin{equation}
\label{eq: single_user}
    \hat{\boldsymbol{\theta}}_{i,t}\approx \mathbf{A}_{i,t}^{-1}\mathbf{X}_{i,t}\mathbf{Y}_{i,t}-\alpha \mathbf{A}_{i,t}^{-1} \sum_{j=1}^n\mathcal{L}_{ij}\mathbf{A}_{j,t}^{-1}\mathbf{X}_{j,t}\mathbf{Y}_{j,t}  
\end{equation}
where $\mathbf{A}_{i,t}=\sum_{\tau \in \mathcal{T}_{i,t}}\mathbf{x}_{\tau}\mathbf{x}_{\tau}^T \in \mathbb{R}^{d \times d}$ is the Gram matrix related to the choices made by user $i$, $\mathcal{T}_{i,t}$ is the set of time at which user $i$ is served up to time $t$, 
and $\mathcal{L}_{ij}$ is the $(i,j)$-th element in $\mathcal{L}$. $\mathbf{X}_{i,t} \in \mathbb{R}^{d \times |\mathcal{T}_{i,t}|}$ is the collection of features of arms that are selected for user $i$ up to time $t$ with $ \{\boldsymbol{x}_{\tau}\}, \tau \in \mathcal{T}_{i,t}$ as columns. $\mathbf{Y}_{i,t} \in \mathbb{R}^{|\mathcal{T}_{i,t}|}$ is the collection of payoffs associated with user $i$ up to time $t$, whose elements are $\{y_\tau\}, \tau \in \mathcal{T}_{i,t}$.
\end{lemma}
\begin{proof}
See Appendix A.
\end{proof}


\subsection{Construction of Confidence Set}
For the agent to balance exploration and exploration in sequential decisions, we need to quantify the uncertainty over the estimation of
$\hat{\boldsymbol{\theta}}_{i,t}$. This is possible  by
defining a confidence set around $\hat{\boldsymbol{\theta}}_{i,t}$ based on Mahalanobis distance using its precision matrix, 
as commonly adopted in bandit literature \cite{lattimore2018bandit}. Let  $\boldsymbol{\Lambda}_{i,t}\in \mathbb{R}^{d\times d}$ be the precision matrix of $\hat{\boldsymbol{\theta}}_{i,t}$,   the confidence set is formally defined as
\begin{equation} 
\label{eq: confidence_set}
\mathcal{C}_{i,t}=\{\boldsymbol{\theta}_{i,t}: ||\hat{\boldsymbol{\theta}}_{i,t}-\boldsymbol{\theta}_i||_{\boldsymbol{\Lambda}_{i,t}}\leq \beta_{i,t}\}
\end{equation}
where $\beta_{i,t}$ is the upper bound of $||\hat{\boldsymbol{\theta}}_{i,t}-\boldsymbol{\theta}_{i}||_{\boldsymbol{\Lambda}_{i,t}}$ which is what we are interested in for the bandit algorithm. With this goal in mind, we seek an expression for   $\boldsymbol{\Lambda}_{i,t}$. Let $\boldsymbol{\Lambda}_{t} \in \mathbb{R}^{nd \times nd}$ denote the precision matrix of $vec(\hat{\boldsymbol{\Theta}}_t)\in \mathbb{R}^{nd}$, where $\boldsymbol{\Lambda}_{i,t} \in \mathbb{R}^{d \times d}$ is the $i$-th block matrix along the diagonal of $\boldsymbol{\Lambda}_t$. Defining the precision matrix of $vec(\hat{\boldsymbol{\Theta}}_t)\in \mathbb{R}^{nd}$ as
\begin{equation}
\label{eq: precision}
    \boldsymbol{\Lambda}_t=\mathbf{M}_t\mathbf{A}_t^{-1}\mathbf{M}_t
\end{equation}
with $\mathbf{A}_t=\mathbf{\Phi}_t\mathbf{\Phi}_t^T$, $\boldsymbol{\mathcal{L}}_{\otimes}=\boldsymbol{\mathcal{L}\otimes \mathbf{I}}$, and $\mathbf{M}_t=\mathbf{A}_t+\alpha \boldsymbol{\mathcal{L}}_{\otimes}$, we have  
\begin{equation}
\label{eq: single_precision}
    \boldsymbol{\Lambda}_{i,t}=\mathbf{A}_{i,t}+2\alpha \mathcal{L}_{ii}\mathbf{I}+\alpha^2\sum_{j=1 }^n\mathcal{L}_{ij}^2 \mathbf{A}_{j,t}^{-1}
\end{equation}
where $\mathbf{A}_{i,t}$ and $\mathbf{A}_{j,t}$ are defined in Lemma \ref{lemma: single_solution},
and $\mathcal{L}_{ij}$ is the $(i,j)$-th element in $\mathcal{L}$. A detailed derivation of Eq.~\ref{eq: single_precision} is presented in Appendix B and Appendix C. Given Eq.~\ref{eq: single_precision}, we can upper bound the size of the confidence set, which provides the value of $\beta_{i,t}$. 
\begin{lemma}
\label{lemma: confidence_bound}
Let $\mathbf{V}_{i,t}=\mathbf{A}_{i,t}+\alpha \mathcal{L}_{ii} \mathbf{I}$,  
and $\mathbf{I}\in \mathbb{R}^{d \times d}$ is the identity matrix. Given a scalar $\delta \in [0,1]$, and by defining $\mathbf{\Delta}_i=\sum_{j=1}^n \mathcal{L}_{ij}\boldsymbol{\theta}_j$, the size of the confidence set defined in Eq.~\ref{eq: confidence_set}
is upper bounded with probability $1-\delta$ by $\beta_{i,t}$:

\begin{equation}
\label{eq: beta}
\begin{split}
    \beta_{i,t}
    &=\sigma \sqrt{2\log \frac{|\mathbf{V}_{i,t}|^{1/2}}{\delta|\alpha \mathbf{I}|^{1/2}}}+\sqrt{\alpha}||\boldsymbol{\Delta}_i||_2
\end{split}
\end{equation}
\end{lemma}
\begin{proof}
See Appendix D.
\end{proof}
It is worth mentioning that the bound $\beta_{i,t}$ depends on  $\mathbf{\Delta}_i$, which reflects information contained in the graph structure. In the case of random-walk Laplacian in which   $\mathcal{L}_{ii}=1$ and $\sum_{j\neq i}-\mathcal{L}_{ij}=1$, $\boldsymbol{\Delta}_i= \boldsymbol{\theta}_i-(\sum_{j\neq i}-\mathcal{L}_{ij}\boldsymbol{\theta}_j)$ measures the difference between user feature $\boldsymbol{\theta}_i$ and the weighted average of its neighbors.  To show the effect of graph structure on $\boldsymbol{\theta}_i$, we consider two extreme cases: 
\textit{a)} an empty graph\footnote{For isolated node, we set $\mathcal{L}_{ii}=1$.}, i.e., $\mathcal{L}_{ii}=1$ and $\mathcal{L}_{ij}=0$. In this case, $\boldsymbol{\Delta}_i=\boldsymbol{\theta}_i$, which recovers the confidence set used in \textbf{LinUCB} \cite{abbasi2011improved}; 
\textit{b)} a fully connected graph with $W_{ij}=1$ which means $\mathcal{L}_{ii}=1$, $\mathcal{L}_{ij}=\frac{1}{n-1}$ and $\boldsymbol{\theta}_i=\boldsymbol{\theta}_j$. In this case, $\boldsymbol{\Delta}_i=\boldsymbol{\theta}_i-\frac{n-1}{n-1}\boldsymbol{\theta}_i=\mathbf{0}$, leading to a much lower bound than the one in case \textit{a)} and with no graph structure to be exploited.  In between, $\boldsymbol{\Delta}_i$ depends on the similarity between $\boldsymbol{\theta}_i$ and its neighbors $\boldsymbol{\theta}_j, j \neq i$.  In general, the smoother the signal is on the graph (in the sense of a small Laplacian quadratic in Eq.~\ref{eq: smoothness}), the lower the $||\boldsymbol{\Delta}_i||_2$. This has been empirically shown in Figure \ref{fig: keys_terms_A}, where we depict    $||\boldsymbol{\Delta}_i||_2$ as a function of the level of smoothness quantified by $tr(\mathbf{\Theta}^T\boldsymbol{\mathcal{L}}\mathbf{\Theta})$ where smaller value means smoother $\mathbf{\Theta}$ over the graph.

Now that we have introduced   $\boldsymbol{\Delta}_i$,   
we are  ready to motivate the choice of the random-walk graph Laplacian  instead of other commonly used graph Laplacians.
\begin{remark}
\label{remark: motivation_of_laplacian}
The two unique properties $\mathcal{L}_{ii}=1$ and $\sum_{j\neq i}-\mathcal{L}_{ij}=1$ of the random-walk graph Laplacian $\boldsymbol{\mathcal{L}}$ ensure a bounded regret and lower regret with more similar users. The same cannot be guaranteed with the combinatorial or normalized Laplacian. 
\end{remark}
To see this more clearly, if the combinatorial Laplacian $\mathbf{L}=\mathbf{D}-\mathbf{W}$ is used in the Laplacian-regularized estimator (Eq.~\ref{eq: objective_function}), the term $\boldsymbol{\Delta}_i$ becomes $\boldsymbol{\Delta}_i=D_{ii}(\boldsymbol{\theta}_i+\sum_{j\neq i}\frac{W_{ij}}{D_{ii}}\boldsymbol{\theta}_j)$. This term scales linearly with $D_{ii}$, the degree of each user $i$,  resulting  in a regret that also scales with $D_{ii}$ and may become rather large for densely connected graphs.
On the other hand, if the symmetric normalized Laplacian $\Tilde{\mathbf{L}}=\mathbf{D}^{-1}\mathbf{L}\mathbf{D}^{-1}$ is used in Eq.~\ref{eq: objective_function}, we have $\sum_{j\neq i}-\Tilde{L}_{ij}\neq 1$. It follows that $\sum_{j\neq i}-\Tilde{L}_{ij}\boldsymbol{\boldsymbol{\theta}}_j$ will not be a convex combination of $\boldsymbol{\theta}_j, j\neq i$, which means that there is no guarantee for   $\boldsymbol{\Delta}_i$ to be located inside the Euclidean ball defined by $||\boldsymbol{\theta}_i||_2\leq 1, \forall i \in [1,...,n]$, leading to an unbounded regret. By contrast, the two unique properties $\mathcal{L}_{ii}=1$ and $\sum_{j\neq i}-\mathcal{L}_{ij}$ of the random-walk normlaized Laplaican $\boldsymbol{\mathcal{L}}$ ensure a bounded regret and less regret if users are similar.

\section{Algorithms}
We  now introduce our proposed  \textbf{GraphUCB} bandit algorithm, sketched in \textbf{Algorithm 1} and based on the principle of \textit{optimism in face of uncertainty}:
\begin{equation}
\label{eq: arm_selection}
\begin{split}
    \mathbf{x}_t
    &=\arg \max_{\mathbf{x}\in \mathcal{D}}\bigg(\mathbf{x}^T\hat{\boldsymbol{\theta}}_{i,t}+\beta_{i,t}||\mathbf{x}||_{\boldsymbol{\Lambda}_{i,t}^{-1}}\bigg)
\end{split}
\end{equation}

\begin{algorithm}[t]
\SetKwInOut{Input}{Input}\SetKwInOut{Output}{Output}
\Input{ $\alpha$, $T$, $\boldsymbol{\mathcal{L}}$, $\delta$}
\textbf{Initialization~~~:} For any $i \in \{1,2,...,n\}$ $\hat{\boldsymbol{\theta}}_{0,i}=\mathbf{0}\in \mathbb{R}^d$,~$\boldsymbol{\Lambda}_{0,i}=\mathbf{0}\in \mathbb{R}^{d \times d}$, $\mathbf{A}_{0,i}=\mathbf{0} \in \mathbb{R}^{d \times d}$, $\beta_{i,t}=0$.
\BlankLine
\For{$t \in [1,T]$}{
User index $i$ is selected
\begin{enumerate}
    \item $\mathbf{A}_{i,t} \gets \mathbf{A}_{i,t-1}+\mathbf{x}_{t-1}\mathbf{x}_{t-1}^T$.
    \item $\mathbf{A}_{j,t}\gets \mathbf{A}_{j,t-1}$, $\forall j\neq i$.
    \item Update $\boldsymbol{\Lambda}_{i,t}$ via Eq. \ref{eq: single_precision}.
    \item Select $\mathbf{x}_{t}$ via Eq. \ref{eq: arm_selection}\\
    where $\beta_{i,t}$ is defined in Eq. \ref{eq: beta} \\
    \item Receive the payoff $y_{t}$ 
    \item Update $\hat{\mathbf{\Theta}}_{t}$ via Eq. \ref{eq: objective_function} 
    \end{enumerate}
}
\caption{\textbf{GraphUCB}}
\label{algorithm:g-ucb}
\end{algorithm}

\textbf{GraphUCB} is designed based on the Laplacian regularized estimator Eq.~\ref{eq: objective_function} and the arm selection principle in Eq.~\ref{eq: arm_selection}. Formally, at each time $t$, an user index $i$ is selected randomly from the user set $\mathcal{U}$. The algorithm first updates $\boldsymbol{\Lambda}_{i,t}$ based on the Gram matrix $\mathbf{A}_{i,t}$ and $\mathbf{A}_{j,t}$. Then, it selects the arm $\mathbf{x}_{t}$ from the arm set $\mathcal{D}$ following Eq.~\ref{eq: arm_selection}. Upon receiving the instantaneous payoff $y_{t}$, it updates the features of all users $\hat{\boldsymbol{\Theta}}_{t}$ by solving Eq.~\ref{eq: objective_function}. The process then continues to time $t+1$, and is repeated until $T$. In  Eq.~\ref{eq: beta} the $\boldsymbol{\Delta}_i$ is based on the unknown ground-truth $\boldsymbol{\theta}_i$ and $\boldsymbol{\theta}_j$. In practice, it is replaced by its empirical counterpart $\hat{\boldsymbol{\Delta}}_i=\sum_{i=1}^n\mathcal{L}_{ij}\hat{\boldsymbol{\theta}}_{i,t}$ where $\hat{\boldsymbol{\theta}}_{i,t}$ is the $i$-th row of $\hat{\mathbf{\Theta}}_t$.


One limitation of \textbf{GraphUCB} is its high computational complexity. Specifically, in solving Eq.~\ref{eq: objective_function}, the running time is dominated by the inversion $(\boldsymbol{\Phi}_t\boldsymbol{\Phi}_t^T+\alpha \boldsymbol{\mathcal{L}}\otimes \mathbf{I})^{-1}$, which is in the order $\mathcal{O}(n^2d^2)$. This is impractical when the user number $n$ is large. 
Recall that in the learning setting, at each time $t$, only one user is selected. Thus, it suffices to only update $\hat{\boldsymbol{\theta}}_{i,t}$ (i.e., a local rather than global update). Therefore, we propose to make use of Lemma \ref{lemma: single_solution} instead of Eq.~\ref{eq: objective_function} to estimate $\hat{\boldsymbol{\theta}}_{i,t}$, which results in a significant reduction in computational complexity. Clearly, the complexity of the approximation in Lemma \ref{lemma: single_solution} is dominated by the inversion of $(\mathbf{A}_{i,t}+\alpha \mathcal{L}_{ii}\mathbf{I})^{-1}$ or $\mathbf{A}_{j,t}^{-1}$, which is in the order  $\mathcal{O}(d^2)$. Since the approximation involves $n$ such inversions, the total complexity is $\mathcal{O}(n d^2)$, i.e., it scales linearly  (rather than quadratically) with $n$.

By using Lemma~\ref{lemma: single_solution}, we therefore propose a second algorithm \textbf{GraphUCB-Local} serving as a simplified version of \textbf{GraphUCB}. The only difference lies in the number of users updated per round. \textbf{GraphUCB} updates all users $\hat{\boldsymbol{\Theta}}_t$ via Eq.~\ref{eq: objective_function} (closed-form solution), while \textbf{GraphUCB-Local} only updates one user $\hat{\boldsymbol{\theta}}_{i,t}$ via Lemma \ref{lemma: single_solution} (approximated solution). Pseudocode of \textbf{GraphUCB-Local} is presented in Appendix E.

\section{Analysis}
Before   providing the finite-time analysis on the proposed algorithms, we define 
\begin{equation}
\label{eq:psi}
\Psi_{i,T}=\frac{\sum_{\tau \in \mathcal{T}_{i,T}}||\mathbf{x}_{\tau}||^2_{\boldsymbol{\Lambda}_{i,\tau}^{-1}}}{\sum_{\tau \in \mathcal{T}_{i,T}}||\mathbf{x}_{\tau}||^2_{\mathbf{V}_{i,\tau}^{-1}}}    
\end{equation}
where $\mathcal{T}_{i,T}$ is the set of time steps in which user $i$ is served up to time $T$, $\mathbf{A}_{i,\tau}=\sum_{\ell \in \mathcal{T}_{i,\tau}}\boldsymbol{x}_\ell \boldsymbol{x}_{\ell}^T$, $\mathbf{V}_{i,\tau}=\mathbf{A}_{i,\tau}+\alpha \mathcal{L}_{ii}\mathbf{I}$ and $\boldsymbol{\Lambda}_{i,\tau}$ is defined as in Eq.~\ref{eq: single_precision}.
 Moreover,  $||\mathbf{x}_\tau||^2_{\boldsymbol{\Lambda}_{i,\tau}^{-1}}$ and $||\mathbf{x}_\tau||^2_{\mathbf{V}_{i,\tau}^{-1}}$ quantify the variance of predicted payoff $\hat{y}_\tau=\hat{\boldsymbol{\theta}}_{i,\tau}^T\mathbf{x}_{\tau}$ in the cases where the graph structured is exploited or ignored, respectively.

\begin{figure}[t]
    \centering
    \subfigure[][]{
        \includegraphics[width=1.5in]{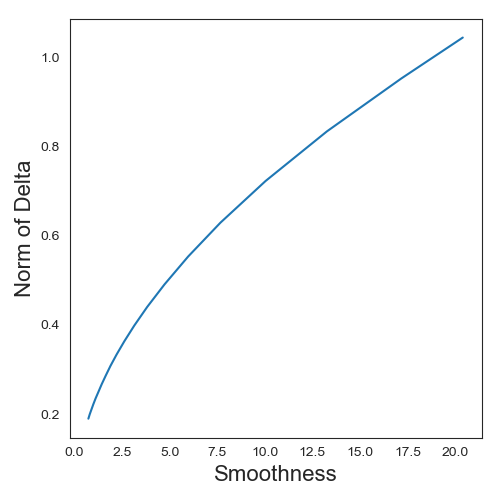}
        \label{fig: keys_terms_A}}
    \subfigure[][]{
        \includegraphics[width=1.5in]{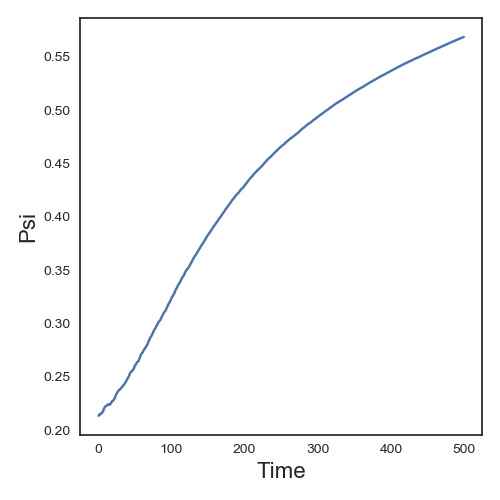}
        \label{fig: keys_terms_B}}
    \vspace{-0.3cm}
\caption{(a) $||\boldsymbol{\Delta}_i||_2$ vs.  smoothness ($tr(\mathbf{\Theta}^T\boldsymbol{\mathcal{L}}\mathbf{\Theta})$), (b) $\Psi_{i,T}$ vs. time.}
\label{fig: keys_terms}
\end{figure}

\begin{lemma}
\label{lemma: psi}
Let $\Psi_{i,T}$ be as defined in Eq.~\ref{eq:psi} and  $||\mathbf{x}_\tau||_2\leq 1$ for any $\tau \leq T$, then 
$$\Psi_{i,T}\in (0,1)$$ 
and as $T \to \infty$, $\Psi_{i,T}\to 1$. This implies 
$$ \sum_{\tau \in \mathcal{T}_{i,T}}||\mathbf{x}_{\tau}||^2_{\boldsymbol{\Lambda}_{i,\tau}^{-1}} \leq {\sum_{\tau \in \mathcal{T}_{i,T}}||\mathbf{x}_{\tau}||^2_{\mathbf{V}_{i,\tau}^{-1}}}\,.$$   
\end{lemma}
Proof on the bound $\Psi_{i,T}$ is provided in Appendix F.  The Lemma  highlights the importance in taking into account the graph structure in the payoff estimation, showing that  the uncertainty of $\hat{y}_{\tau}$ is reduced when the graph structure is exploited. This effect diminishes with time: in Fig.~\ref{fig: keys_terms_B}, we see that as more data are collected, the graph-based estimator approaches the estimator in which users parameters are estimated independently (since $\Psi_{i,T}\to 1$).

\subsection{Regret Upper Bound}
We present the cumulative regret upper bounds satisfied by both \textbf{GraphUCB} and \textbf{GraphUCB-Local}.
\begin{theorem}
\label{theorem: regret_upper_bound}
$\Psi_{i,T}$ is defined in Eq.~\ref{eq:psi},
$\boldsymbol{\Lambda}_{i,T}$ defined in Eq.~\ref{eq: single_precision} and  $\boldsymbol{\Delta}_i=\sum_{j=1}^n \mathcal{L}_{ij}\boldsymbol{\theta}_j$. Without loss of generality, assume  $||\boldsymbol{\theta}_i||_2\leq 1$ for any $i \in \{1,2,...,n\}$ and  $||\mathbf{x}_\tau||_2\leq 1$ for any $\tau \leq T$. Then, for $\delta \in [0,1]$, for any user $i \in \{1,2,...,n\}$ the cumulative regret over time horizon $T$ satisfies the following upper bound with probability $1-\delta$ 
\begin{equation}
\label{eq: regret_upper_bound}
\begin{split}
    R_{i,T}
    &=\sum_{\tau \in \mathcal{T}_{i,T}}r_\tau
    =\mathcal{O}\bigg(\big(\sqrt{d\log(|\mathcal{T}_{i,T}|)}+\sqrt{\alpha} ||\boldsymbol{\Delta}_i||_2\big)\times\\ &\Psi_{i, T}\sqrt{d|\mathcal{T}_{i,T}|\log(|\mathcal{T}_{i,T}|)}\bigg)
\end{split}
\end{equation}
Assuming that users are served uniformly up to time horizon $T$, i.e., $|\mathcal{T}_{i,T}|=T/n$, the network regret (the total cumulative regret experienced by all users) satisfies the following upper bound with probability $1-\delta$:
\begin{equation}
\label{eq: total_regret_upper_bound}
\begin{split}
  R_T
  &=\sum_{i=1}^nR_{i,T}=\sum_{i=1}^n\Tilde{\mathcal{O}}\bigg(\Psi_{i,T}d\sqrt{T/n}\bigg)\\
  &=\Tilde{\mathcal{O}}\bigg( d\sqrt{Tn}\max_{i\in \mathcal{U}}\Psi_{i,T}\bigg) 
\end{split}
\end{equation}
\end{theorem}
\begin{proof}
Appendix G.
\end{proof}

\begin{remark}
\label{remark: gucb_and_gucbsim}
Both \textbf{GraphUCB} and \textbf{GraphUCB-Local} satisfy Theorem \ref{theorem: regret_upper_bound}. 
\end{remark}
The regret upper bound in Theorem \ref{theorem: regret_upper_bound} is derived based on \textbf{GraphUCB-Local} algorithm (Appendix G). Due to the approximation error introduced in the Taylor expansion, the regret of \textbf{GraphUCB-Local} is worse than that of \textbf{GraphUCB}. Therefore, Theorem \ref{theorem: regret_upper_bound} is also an upper bound of \textbf{GraphUCB}.

\subsection{Comparison with \textbf{LinUCB} and \textbf{Gob.Lin}}
Under the same setting, the single-user regret upper bound of \textbf{LinUCB} \cite{li2010contextual} is 
\begin{equation}
\label{eq: linucb_regret_upper_bound}
\begin{split}
    R_{i,T}
    &=\mathcal{O}\bigg(\big(\sqrt{d\log(|\mathcal{T}_{i,T}|)}+\sqrt{\alpha}||\boldsymbol{\theta}_i||_2\big)\times \\
    &\sqrt{d|\mathcal{T}_{i,T}|\log(|\mathcal{T}_{i,T}|)}\bigg)\\
\end{split}
\end{equation}
Since $||\boldsymbol{\Delta}_i||_2\leq ||\boldsymbol{\theta}_i||_2$ and $\Psi_{i,T}\in (0,1)$ (Lemma \ref{lemma: confidence_bound} and Lemma \ref{lemma: psi}),  \textbf{GraphUCB} (and \textbf{GraphUCB-Local}) leads to a lower regret $-$Eq.~\ref{eq: regret_upper_bound}$-$   than \textbf{LinUCB} $-$Eq.~\ref{eq: linucb_regret_upper_bound}. 

The cumulative regret experienced by all users in  \textbf{Gob.Lin} \cite{cesa2013gang} is upper bounded
by
\begin{equation}
\label{eq: gob_regret}
\begin{split}
    R_T
    &=4\sqrt{T(\sigma^2\log \frac{|\mathbf{M}_t|}{\delta}+L(\boldsymbol{\theta})\log |\mathbf{M}_t|}=\Tilde{\mathcal{O}}(nd\sqrt{T})
\end{split}
\end{equation}
where $L(\boldsymbol{\theta})=\sum_{i=1}^n||\boldsymbol{\theta}_i||_2+\sum_{(i,j)\in E}||\boldsymbol{\theta}_i-\boldsymbol{\theta}_j||_2$.
Clearly, the cumulative regret achieved by \textbf{GraphUCB} in Eq.~\ref{eq: total_regret_upper_bound}
is in the order of $\Tilde{\mathcal{O}}(\sqrt{n})$ which is better than that in Eq.~\ref{eq: gob_regret}.
This is mainly due to the different UCBs used in these algorithms. Specifically, \textbf{Gob.Lin} proposed the following bound.
\begin{equation}
\label{eq: gob_beta}
    \beta_{t}
    =\sigma \sqrt{\log \frac{|\mathbf{M}_t|}{\delta}+L(\boldsymbol{\theta})}= \Tilde{\mathcal{O}}(\sqrt{nd})
\end{equation}
while we propose a  lower single-user bound $\beta_{i,t} =\Tilde{\mathcal{O}}(\sqrt{d})$ (Eq.~\ref{eq: beta}).
As described in Remark \ref{remark: motivation_of_laplacian}, the bound in Eq.~\ref{eq: gob_beta} grows with the degree of the network, which in the bound is hidden in  $L(\boldsymbol{\theta})$.
We emphasize that in practice \textbf{GOB.Lin} could perform much better than its regret upper bound Eq.~\ref{eq: gob_regret}, if $\beta_t$ defined in Eq.~\ref{eq: gob_beta} is replaced by $\lambda \sqrt{log(t+1)}$ where $\lambda$ is a tunable parameter. This is exactly what the authors of $\textbf{Gob.Lin}$ did in their original paper when reporting empirical results. We follow the same trick when implementing \textbf{Gob.Lin} in our experiments, and show that \textbf{GraphUCB} still achieves better performance empirically.

\section{Experiment Results}
We evaluate the proposed algorithms and compare them to \textbf{LinUCB} (no graph information exploited in the bandit), \textbf{Gob.Lin} (graph exploited in the features estimation) and \textbf{CLUB} (graph exploited to cluster users). 
All results reported are averaged across 20 runs.
In all experiments, we set confidence probability parameter $\delta=0.01$, noise variance $\sigma=0.01$, and regularization parameter $\alpha=1$. For \textbf{Gob.Lin}, we use $\beta_{i,t}=\lambda \sqrt{\log(t+1)}$, and  $\lambda$ is set using the best value in range $[0,1]$. For \textbf{CLUB}, the edge deletion parameter $\alpha_2$ is tuned to its best value.

\subsection{Experiments on Synthetic Data}
In the synthetic simulations, we first generate a graph $\mathcal{G}$ and then generate a smooth $\boldsymbol{\Theta}$ via Eq.~\ref{eq: smooth_signal} which is proposed in \cite{yankelevsky2016dual}: 
\begin{equation}
\label{eq: smooth_signal}
  \boldsymbol{\Theta}=\arg \min_{\boldsymbol{\Theta}\in \mathbb{R}^{n \times d}}||\boldsymbol{\Theta}-\boldsymbol{\Theta}_0||_F^2+\gamma tr(\boldsymbol{\Theta}^T\mathcal{L}\boldsymbol{\Theta})
\end{equation}
where $\mathbf{\Theta_0} \in \mathbb{R}^{n \times d}$ is a randomly initialized matrix, and $\boldsymbol{\mathcal{L}}$ is the random-walk graph Laplacian of $\mathcal{G}$. The second term in Eq.~\ref{eq: smooth_signal} promotes the smoothness of $\boldsymbol{\Theta}$: the larger the $\gamma$, the smoother the $\boldsymbol{\Theta}$ over the graph\footnote{The regularization parameter $\gamma$ in Eq.~\ref{eq: smooth_signal} is used to generate a smooth function in the synthetic settings, while the parameter $\alpha$ in Eq.~\ref{eq: objective_function} is used in the bandit algorithm to infer the smooth prior when estimating user features.}.
In all experiments, $n=20, d=5$.
To simulate $\mathcal{G}$, we follow two random graph models commonly used in the network science community: 1) Radial basis function (\textit{RBF}) model, a weighted fully connected graph, with edge weights $W_{ij}=exp(-\rho||\boldsymbol{\theta_i}-\boldsymbol{\theta_j}||^2)$; 2) Erd\H{o}s R\'enyi (ER) model, an unweighted graph, in which each edge is generated independently and randomly with probability $p$. 
\begin{figure}[t]
    \centering
    \subfigure[][RBF (s=0.5)]{
        \includegraphics[width=1.5in]{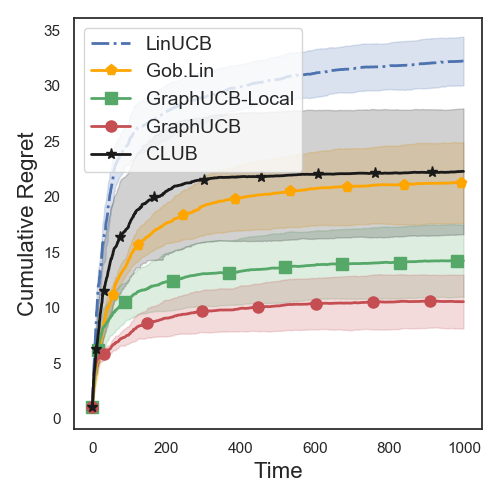}
        }
    \subfigure[][ER (p=0.4)]{
        \includegraphics[width=1.5in]{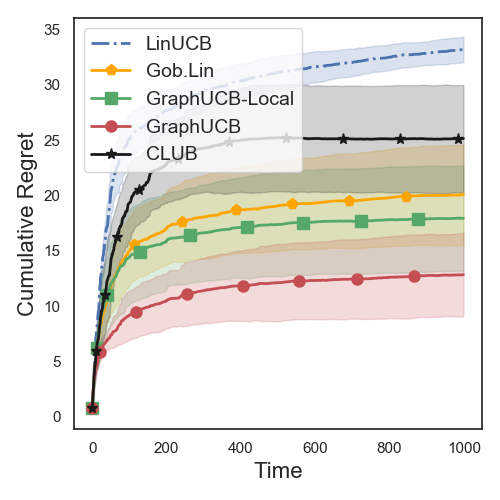}
        }
    \vspace{-0.3cm}
\caption{Cumulative regret vs. time for different type of graphs (ER and RBF) consistently generated with the same level of smoothness and sparsity between graphs.
}
\label{fig: performance_on_graph_structures_weighted}
\end{figure}

\begin{figure}[t]
    \centering
    \subfigure[][]{
        \includegraphics[width=1.5in]{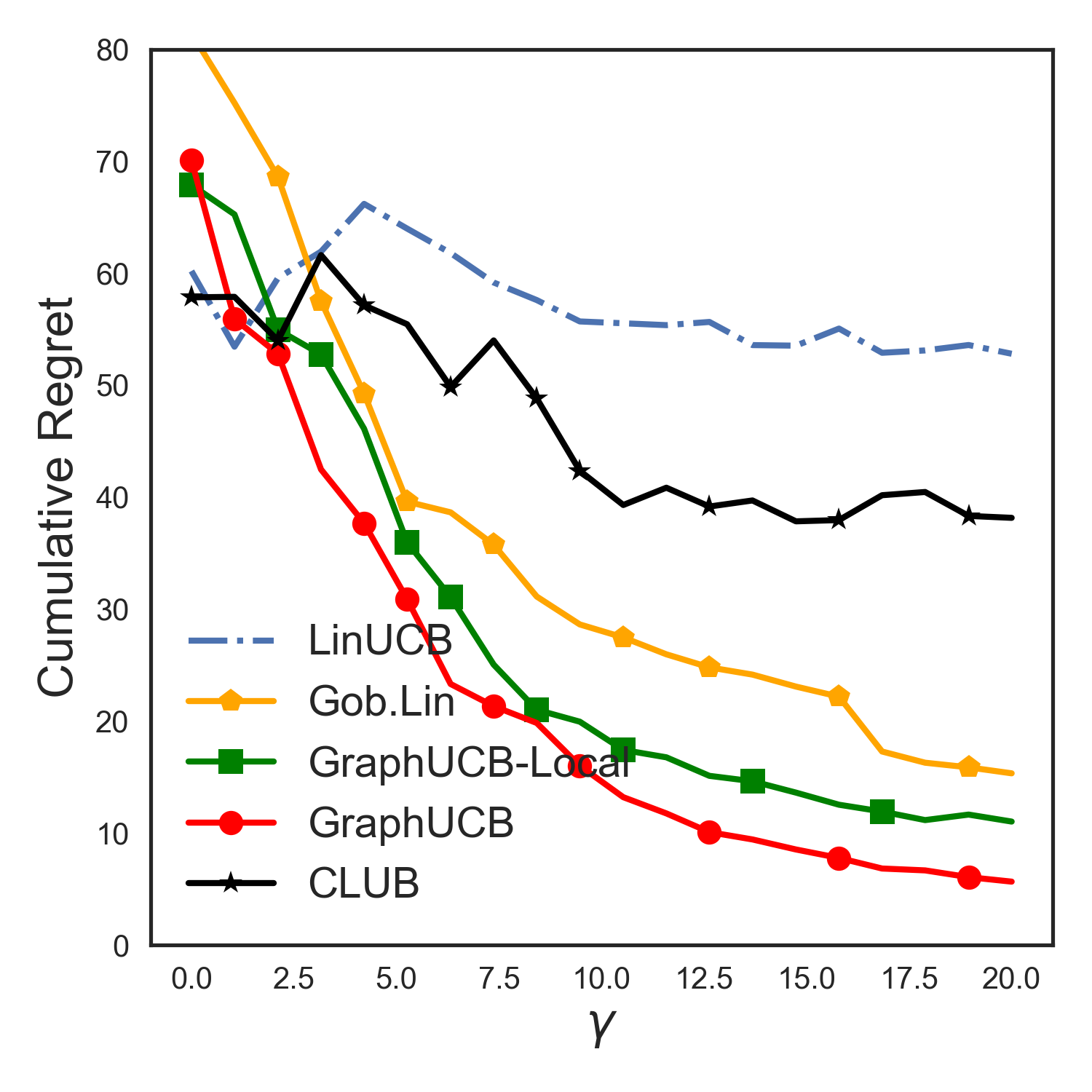}
        }
    \subfigure[][ ]{
        \includegraphics[width=1.5in]{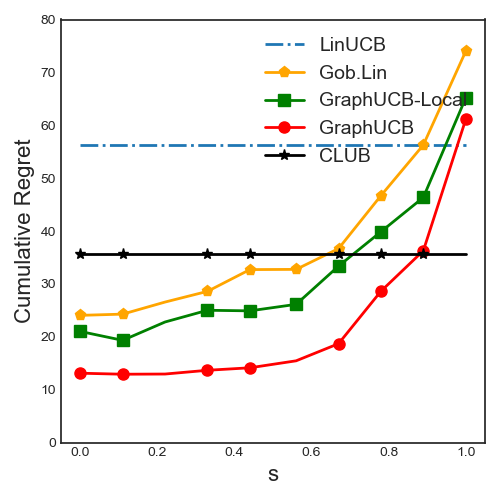}
        }
    \vspace{-0.3cm}
\caption{Cumulative regret for RBF graphs with different level of smoothness (a) and sparsity (b).}
\label{fig: performance_on_graph_properties}
\end{figure}

In Figure~\ref{fig: performance_on_graph_structures_weighted}, we depict the cumulative per-user regret as a function of time for both RBF and ER graphs for both our proposed algorithms and competitors. The regret is averaged over all users and over all runs.  Under all graph models, \textbf{GraphUCB} outperforms its competitors consistently with a large margin. Also \textbf{GraphUCB-Local} consistently outperform competitor algorithms, with however slightly degradated performance  with respect to   \textbf{GraphUCB}. This is due to the approximation introduced by Eq.~\ref{eq: single_user}. \textbf{Gob.Lin} is a close runner since it is also based on the Laplacian regularized estimator, but it performs worse than the proposed algorithms, as already explained in the previous section.  \textbf{CLUB} performs relative poor since there is no clear clusters in the graph. Nevertheless, it still outperforms \textbf{LinUCB} by grouping users into clusters in the early stage which speeds up the learning process. It is worth noting that the two subfigures depict the same algorithm for two different graph models (RBF and ER) with the same level of smoothness and sparsity. The trend of the cumulative regret is the same, highlighting that the algorithm is not affected by the graph model. This behavior is reinforced in Appendix J where we provide further results.

We are now interested in evaluating the performance of the proposed algorithms against different graph topologies, by varying signal smoothness and sparsity of graph (edge density)   as follows 
\newline
\textbf{Smoothness [$\gamma$]:} We first generate a \textit{RBF} graph. To control the smoothness, we vary $\gamma \in [0,10]$.
\newline
\textbf{Sparsity [$s$]:}
We first generate a \textit{RBF} graph, then generate a smooth $\mathbf{\Theta}$ via Eq.~\ref{eq: smooth_signal}.
To control the sparsity, we set a threshold $s \in [0,1]$ on edge weights $W_{ij}$ such that $W_{ij}$ less than $s$ are removed. 

Figure~\ref{fig: performance_on_graph_properties}, depicts the cumulative regret for different level of smoothness and sparsity of $\mathcal{G}$. \textbf{GraphUCB} and \textbf{GraphUCB-Local} show similar patterns (with \textbf{GraphUCB-Local} leading to higher regret due to the already commented approximation): \textit{(i)} the smoother $\mathbf{\Theta}$ the lower is regret, which is consistent with the Laplacian-regualrized estimator Eq.~\ref{eq: objective_function}; \textit{(ii)} denser graphs lead to lower regret since more connectivity provides more graph information which speeds up the learning process. 
\newline
\subsection{Experiments on Real-World Data}
We then carry out experiments on two real-world datasets : \textbf{Movielens} \cite{lam2006movielens} and  \textbf{Netflix} \cite{bennett2007netflix}. We follow the data pre-processing steps in \cite{valko2014spectral}, described in details in Appendix K.
\begin{figure}[t]
    \centering
    \subfigure[][MovieLens]{
        \includegraphics[width=1.5in]{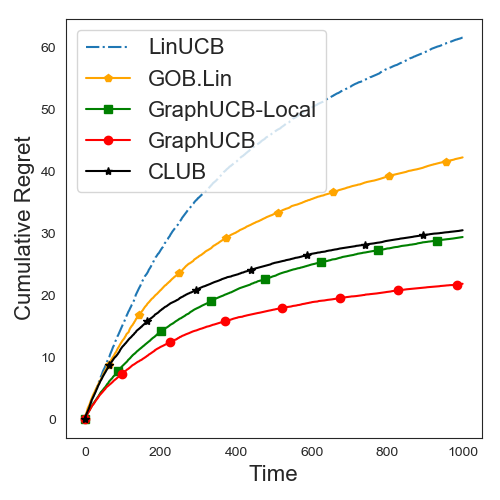}
        }
    \subfigure[][Netflix]{
        \includegraphics[width=1.5in]{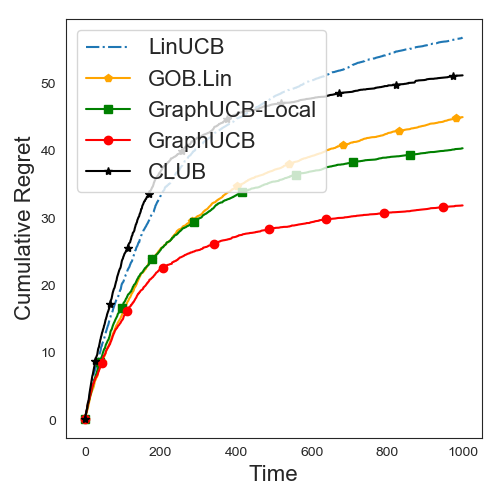}
        }
    \vspace{-0.3cm}
\caption{Performance on Real-World data.}
\label{fig: performance_on_real_data}
\end{figure}
The cumulative regret over time is depicted in  Figure~\ref{fig: performance_on_real_data} for both datasets. Both \textbf{GraphUCB} and \textbf{GraphUCB-Local} outperform baseline algorithms in all cases. Similarly to the synthetic experiments, \textbf{LinUCB} performs poorly, while \textbf{Gob.Lin} shows a regret behavior more similar to the proposed algorithms. In the case of \textbf{Movielens}, \textbf{CLUB} outperforms \textbf{Gob.Lin}. A close inspection of the data reveals that ratings provided by all users are highly concentrated. It means most users like a few set of movies. This is a good model for the clustering algorithm implemented in  \textbf{CLUB}, hence the gain.

\section{Conclusion}

In this work, we propose a graph-based bandit algorithm \textbf{GraphUCB} and its scalable version \textbf{GraphUCB-Local}, both of which outperform the state-of-art bandit algorithms in terms of cumulative regret. On the theoretical side, we introduce a novel UCB embedding the graph structure in a natural way and show clearly that exploring the graph prior could reduce the cumulative regret. We demonstrate that the graph structure helps reduce the size of confidence set of the estimation of user features and the uncertainty of predicted payoff. As for future research directions, one possibility is to relax the assumption that the user graph is available and infer the graph from the data, ideally in a dynamic fashion.

\newpage
\bibliography{main}

\clearpage 

\newpage
\section*{Appendix A}
\begin{figure}[h]
\begin{center}
\includegraphics[width=2in]{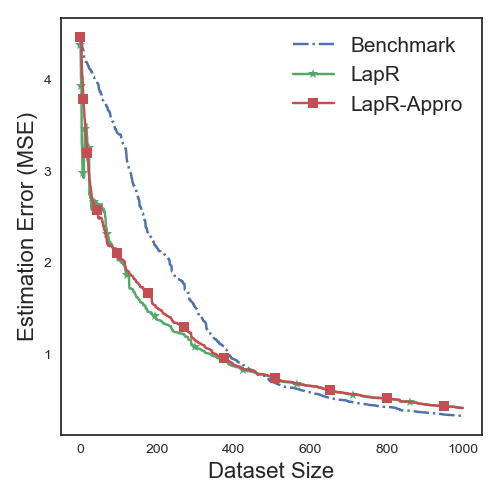}
\end{center}
\caption{Approximation accuracy}
\label{fig: single_user_approximation}
\end{figure}

\textbf{Proof of Lemma \ref{lemma: single_solution}}
\begin{proof}
\begin{equation}
    vec(\hat{\boldsymbol{\Theta}}_t)=(\mathbf{A}_t+\alpha \boldsymbol{\mathcal{L}_{\otimes}})^{-1}\mathbf{\Phi}_t\mathbf{Y}_t
\end{equation}
Let $\mathbf{M}_t=\mathbf{A}_t+\alpha \boldsymbol{\mathcal{L}_{\otimes}}$ and $\mathbf{B}_t=\mathbf{\Phi}_t\mathbf{Y}_t$, then
\begin{equation}
    vec(\hat{\boldsymbol{\Theta}}_t)=\mathbf{M}_t^{-1}\mathbf{B}_t
\end{equation}
Express $\mathbf{M}_t$ and $\mathbf{B}_t$ in partitioned form. For instance,  if $n=2$
\begin{equation}
    \mathbf{M}_t^{-1}=
    \begin{bmatrix}
    \mathbf{M}_{11,t}^{-1}&\mathbf{M}^{-1}_{12,t}\\
    \mathbf{M}^{-1}_{21,t} &\mathbf{M}^{-1}_{22,t}
    \end{bmatrix} \ 
    \mathbf{B}_t=
    \begin{bmatrix}
    \mathbf{B}_{1,t}\\
    \mathbf{B}_{2,t}
    \end{bmatrix}
\end{equation}
Then 
\begin{equation}
    \hat{\boldsymbol{\theta}}_{i,t}=\sum_{j=1}^2\mathbf{M}_{ij,t}^{-1}\mathbf{B}_{j,t}
\end{equation}
In general case, $n\geq 2$,
\begin{equation}
\hat{\boldsymbol{\theta}}_{i,t}=\sum_{j=1}^n \mathbf{M}_{ij,t}^{-1}\mathbf{B}_{j,t}
\end{equation}
To obtain the expression of $\hat{\boldsymbol{\theta}}_{i,t}$, we need the close form of $\mathbf{M}^{-1}_{ij,t}$. 
Given $\mathbf{M}_t=\mathbf{A}_{t}+\alpha \boldsymbol{\mathcal{L}}_{\otimes}$ we have
\begin{equation}
\begin{split}
    \mathbf{M}_t^{-1}
    &=(\mathbf{A}_t+\alpha \boldsymbol{\mathcal{L}}_{\otimes})^{-1}\\
    &=(\mathbf{A}_{t}\mathbf{A}_t^{-1}\mathbf{A}_t+\alpha\boldsymbol{\mathcal{L}}_{\otimes}\mathbf{A}^{-1}_t\mathbf{A}_t)^{-1}\\
    &=\bigg((\mathbf{I}+\alpha \boldsymbol{\mathcal{L}}_{\otimes}\mathbf{A}_t^{-1})\mathbf{A}_t\bigg)^{-1}\\
    &=\mathbf{A}_t^{-1}(\mathbf{I}+\alpha \boldsymbol{\mathcal{L}}_{\otimes}\mathbf{A}_t^{-1})^{-1}
\end{split}
\end{equation}
This can be rewritten using Taylor expansion
\begin{equation*}
    \mathbf{A}_t^{-1}(\mathbf{I}+\alpha \boldsymbol{\mathcal{L}}_{\otimes}\mathbf{A}_t^{-1})^{-1}
\end{equation*}
\begin{equation*}
    =\mathbf{A}_t^{-1}\bigg(\mathbf{I}-\alpha \boldsymbol{\mathcal{L}}_{\otimes}\mathbf{A}_t^{-1}+(\alpha \boldsymbol{\mathcal{L}}_{\otimes}\mathbf{A}_t^{-1})^2-... \bigg)
\end{equation*}
\begin{equation}
\label{eq: taylor_expansion}
    \approx \mathbf{A}_t^{-1}-\alpha\mathbf{A}^{-1}_t \boldsymbol{\mathcal{L}}_{\otimes}\mathbf{A}_t^{-1}
\end{equation}
The last step keeps the first two terms only.
So, 
\begin{equation}
\label{eq: M_inverse_proof}
    \mathbf{M}^{-1}_t\approx \mathbf{A}_t^{-1}-\alpha \mathbf{A}_t^{-1}\boldsymbol{\mathcal{L}}_{\otimes}\mathbf{A}^{-1}_t
\end{equation}
To obtain the expression of $\mathbf{M}^{-1}_{ij,t}$, we need the partitioned form of $\mathbf{A}^{-1}_t$ and $\boldsymbol{\mathcal{L}}_{\otimes}$.

Since $\mathbf{A}_t$ is a block diagonal matrix, the inversion of it is the inversion of its block matrix. For instance, if $n=2$
\begin{equation}
    \mathbf{A}_t^{-1}=
    \begin{bmatrix}
    \mathbf{A}_{1,t}^{-1}&0\\
    0&\mathbf{A}_{2,t}^{-1}
    \end{bmatrix}
\end{equation}
and 
\begin{equation}
    \boldsymbol{\mathcal{L}}_{\otimes}=
    \begin{bmatrix}
    \mathcal{L}_{11}\mathbf{I}&\mathcal{L}_{12}\mathbf{I}\\
    \mathcal{L}_{21}\mathbf{I}& \mathcal{L}_{22}\mathbf{I}
    \end{bmatrix}
\end{equation}
Given Eq.~\ref{eq: M_inverse_proof},
in general case, $n\geq 2$, it is trivial to show
\begin{equation}
    \mathbf{M}^{-1}_{ij, t}\approx
    \begin{cases}
       \mathbf{A}^{-1}_{i,t}-\alpha \mathbf{A}_{i,t}^{-1}\mathcal{L}_{ii}\mathbf{A}_{i,t}^{-1} & when \ i=j \\
        -\alpha \mathbf{A}_{i,t}^{-1}\mathcal{L}_{ij}\mathbf{A}_{j,t}^{-1} & when \ i \neq j
    \end{cases}
\end{equation}
Finally, we have
\begin{equation}
\label{eq: single_user_proof}
\begin{split}
    \hat{\theta}_{i,t}
    &=\sum_{j=1}^n\mathbf{M}_{ij,t}^{-1}\mathbf{B}_{j,t}\\
    &\approx (\mathbf{A}_{i,t}^{-1}-\alpha\mathbf{A}_{i,t}^{-1}\mathcal{L}_{ii}\mathbf{A}_{i,t}^{-1})\mathbf{B}_{i,t}\\
    &-\alpha \mathbf{A}_{i,t}^{-1}\sum_{j \neq i}\mathcal{L}_{ij}\mathbf{A}_{j,t}^{-1}\mathbf{B}_{j,t}\\
    &=\mathbf{A}_{i,t}^{-1}\mathbf{B}_{i,t}-\alpha \mathbf{A}_{i,t}^{-1}\sum_{j=1}^n\mathcal{L}_{ij}\mathbf{A}_{j,t}^{-1}\mathbf{B}_{j,t}
\end{split}
\end{equation}
We claim the approximation introduced in Eq.~\ref{eq: single_user_proof} is small. To see this, tracing back to Eq.~\ref{eq: taylor_expansion} where the higher order terms are dropped. These higher orders terms are negligible when $t$ is large. This is because the Gram matrix $\mathbf{A}_{t}$ grows with larger $t$, in turn $\mathbf{A}_{t}^{-1}$ decreases. In such case, higher order terms would be incomparable with lower order terms. 

To support the tightness of this approximation, we provide an empirical evidence in Figure~\ref{fig: single_user_approximation}. Red curve represents the single-user estimation error of The Laplacian-regularised estimator Eq.~\ref{eq: objective_function}, while green curve is the estimation error of Lemma \ref{lemma: single_solution}.  Blue curve represents the estimation error of ridge regression, which is included as a benchmark. Clearly, Lemma \ref{lemma: single_solution} is a tight approximation of Eq.~\ref{eq: objective_function} and both converge faster than ridge regression due to the smoothness prior. 
\end{proof}

\section*{Appendix B}
\textbf{Proof of Eq.~\ref{eq: precision}}. 
\begin{proof}
Let $\mathbf{A}_t=\mathbf{\Phi}_t\mathbf{\Phi}_t^T$ and $\boldsymbol{\mathcal{L}}_{\otimes}=\boldsymbol{\mathcal{L}\otimes \mathbf{I}}$ and $\mathbf{M}_t=\mathbf{A}_t+\alpha \boldsymbol{\mathcal{L}}_{\otimes}$.  $\boldsymbol{\epsilon}_t=[\eta_1, \eta_2,...,\eta_t]$
\begin{equation}
    vec(\hat{\boldsymbol{\Theta}}_t)=\mathbf{M}_t^{-1}\mathbf{\Phi}_t\mathbf{Y}_t
\end{equation}



The variance $\boldsymbol{\Sigma}_t$ is 
\begin{equation}
\begin{split}
   \boldsymbol{\Sigma}_t=Cov(vec(\hat{\boldsymbol{\Theta}}_t))
   &=Cov(\mathbf{M}_t^{-1}\mathbf{\Phi}_t\mathbf{Y}_t)\\
   &=\mathbf{M}_t^{-1}\mathbf{\Phi}_tCov(\mathbf{Y}_t)\mathbf{\Phi}_t^T\mathbf{M}_t^{-1}\\
   &=\sigma^2\mathbf{M}_t^{-1}\mathbf{A}_t\mathbf{M}_t^{-1}
\end{split}
\end{equation}
where we use $Cov(\mathbf{Y}_t)=\sigma^2\mathbf{I}$ since noise follows $\mathcal{N}(0, \sigma^2)$. 
Therefore, the precision matrix
\begin{equation}
   \boldsymbol{\Lambda}_t=\boldsymbol{\Sigma}_t^{-1}=\frac{1}{\sigma^2}\mathbf{M}_t\mathbf{A}_t^{-1}\mathbf{M}_t 
\end{equation}
For simplicity, we assume $\sigma=1$
\begin{equation}
   \boldsymbol{\Lambda}_t=\boldsymbol{\Sigma}_t^{-1}=\mathbf{M}_t\mathbf{A}_t^{-1}\mathbf{M}_t 
\end{equation}
\end{proof}

\section*{Appendix C}
\textbf{Proof of Eq.~\ref{eq: single_precision}}
\begin{proof}
Recall 
\begin{equation}
    \boldsymbol{\Lambda}_t=\mathbf{M}_t\mathbf{A}_t^{-1}\mathbf{M}_t
\end{equation}
and $\boldsymbol{\Lambda}_{i,t}\in \mathbb{R}^{d \times d}$ is the $i$-th block matrix along the diagonal of $\boldsymbol{\Lambda}_t$. 

To get the expression of $\boldsymbol{\Lambda}_{i,t}$, we need to
express $\mathbf{M}_t$, $\mathbf{A}_t^{-1}$ in partitioned form. For instance, $n=2$
\begin{equation}
    \mathbf{M}_t=
    \begin{bmatrix}
    \mathbf{M}_{11,t} & \mathbf{M}_{12,t}\\
    \mathbf{M}_{21,t} & \mathbf{M}_{22,t}
    \end{bmatrix}
    , \ 
    \mathbf{A}_t^{-1}=
    \begin{bmatrix}
    \mathbf{A}_{1,t}^{-1} &\mathbf{0}\\
    \mathbf{0} & \mathbf{A}_{2,t}^{-1}
    \end{bmatrix}
\end{equation}
In general case $n \geq 2$, it is straightforward to see 
\begin{equation}
\begin{split}
    \boldsymbol{\Lambda}_{i,t}
    &=\sum_{j=1}^n\mathbf{M}_{ij,t}\mathbf{A}_{i,t}^{-1}\mathbf{M}_{ij,t}\\
    &=\bigg(\mathbf{M}_{ii,t}\mathbf{A}_{i,t}^{-1}\mathbf{M}_{ii,t}+\sum_{j\neq i}\mathbf{M}_{ij,t}\mathbf{A}_{j,t}^{-1}\mathbf{M}_{ji,t}  \bigg)
\end{split}
\end{equation}

From $\mathbf{M}_t=\mathbf{A}_t+\alpha \boldsymbol{\mathcal{L}}_{\otimes}$, we know
\begin{equation}
    \mathbf{M}_{ii,t}=\mathbf{A}_{i,t}+\alpha \mathcal{L}_{ii}\mathbf{I}
\end{equation}
\begin{equation}
    \mathbf{M}_{ij,t}=\alpha \mathcal{L}_{ij}\mathbf{I}
\end{equation}
Hence, 
\begin{equation}
    \boldsymbol{\Lambda}_{i,t}=\bigg( \mathbf{A}_{i,t}+2\alpha \mathcal{L}_{ii}\mathbf{I}+\alpha^2\sum_{j=1 }^n\mathcal{L}_{ij}^2 \mathbf{A}_{j,t}^{-1}\bigg)
\end{equation}
\end{proof}

\section*{Appendix D}
\begin{figure}[ht]
\begin{center}
\includegraphics[width=2.5in]{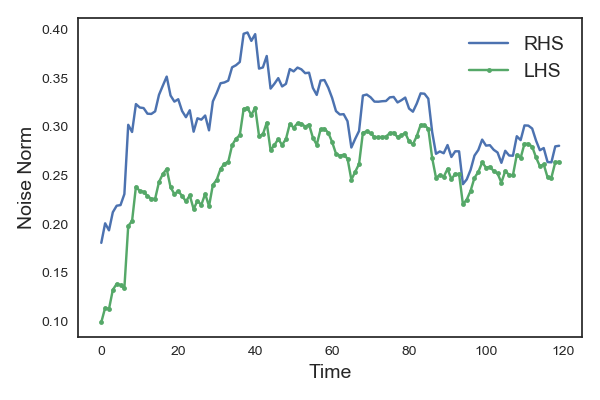}
\end{center}
\caption{Noise term approximation}
\label{fig: noise_term_approximation}
\end{figure}

\textbf{Proof of Lemma \ref{lemma: confidence_bound}}
\begin{proof}
\begin{equation}
    \mathcal{C}_t=\{\boldsymbol{\theta}_i: ||\hat{\boldsymbol{\theta}}_{i,t}-\boldsymbol{\theta}_i||_{\boldsymbol{\Lambda}_{i,t}} \leq \beta_{i,t}\}
\end{equation}
From Lemma \ref{lemma: single_solution}, we have
\begin{equation}
    \hat{\boldsymbol{\theta}}_{i,t}\approx\mathbf{A}_{i,t}^{-1}\mathbf{X}_{i,t}\mathbf{Y}_{i,t}-\alpha \mathbf{A}_{i,t}^{-1}\sum_{j=1}^n \mathcal{L}_{ij}\mathbf{A}_{j,t}^{-1}\mathbf{X}_{j,t}\mathbf{Y}_{j,t}
\end{equation}
Note that $\mathbf{Y}_{i,t}=\mathbf{X}_{i,t}^T\boldsymbol{\theta}_i+\boldsymbol{\epsilon}_{i,t}$ and $\mathbf{Y}_{j,t}=\mathbf{X}_{j,t}^T\boldsymbol{\theta}_j+\boldsymbol{\epsilon}_{j,t}$ where $\boldsymbol{\epsilon}_{i,t}=[\eta_{i,1}, ..., \eta_{i,\mathcal{T}_{i,t}}]$ is the collection of noise associated with user $i$. $\mathcal{T}_{i,t}$ the set of time user $i$ is selected during the time period from 1 to $t$.

Then we have
\begin{equation}
    \begin{split}
        \hat{\boldsymbol{\theta}}_{i,t}
        &=\mathbf{A}_{i,t}^{-1}\mathbf{X}_{i,t}(\mathbf{X}_{i,t}^T\boldsymbol{\theta}_i+\boldsymbol{\epsilon}_{i,t})\\
        &-\alpha \mathbf{A}_{i,t}^{-1}\sum_{j=1 }^n\mathcal{L}_{ij}\mathbf{A}_{j,t}^{-1}\mathbf{X}_{j,t}(\mathbf{X}_{j,t}^T\boldsymbol{\theta}_j+\boldsymbol{\epsilon}_{j,t})\\
        &=\boldsymbol{\theta}_i+\mathbf{A}_{i,t}^{-1}\mathbf{X}_{i,t}\boldsymbol{\epsilon}_{i,t}
        -\alpha \mathbf{A}_{i,t}^{-1}\sum_{j=1 }^n\mathcal{L}_{ij}\boldsymbol{\theta}_j\\
        &-\alpha \mathbf{A}_{i,t}^{-1}\sum_{j \neq i}\mathcal{L}_{ij}\mathbf{A}_{j,t}^{-1}\mathbf{X}_{j,t}\boldsymbol{\epsilon}_{j,t}
    \end{split}
\end{equation}
Hence
\begin{equation}
    \begin{split}
    \hat{\boldsymbol{\theta}}_{i,t}-\boldsymbol{\theta}_i
    &=\mathbf{A}_{i,t}^{-1}\mathbf{X}_{i,t}\boldsymbol{\epsilon}_{i,t}-\alpha \mathbf{A}_{i,t}^{-1}\sum_{j=1}^n \mathcal{L}_{ij}\boldsymbol{\theta}_j\\
    &-\alpha \mathbf{A}_{i,t}^{-1}\sum_{j=1}^n \mathcal{L}_{ij}\mathbf{A}_{j,t}^{-1}\mathbf{X}_{j,t}\boldsymbol{\epsilon}_{j,t}
    \end{split}
\end{equation}
Denote $\boldsymbol{\xi}_{i,t}=\mathbf{X}_{i,t}\boldsymbol{\epsilon}_{i,t}$ and $\mathbf{V}_{i,t}=\mathbf{A}_{i,t}+\alpha \mathcal{L}_{ii}\mathbf{I}$, then, 

\begin{equation}
    \begin{split}
    ||\hat{\boldsymbol{\theta}}_{i,t}-\boldsymbol{\theta}_{i}||_{\boldsymbol{\Lambda}_{i,t}}
    &\leq ||\alpha \mathbf{A}_{i,t}^{-1}\sum_{j=1}^n \mathcal{L}_{ij}\boldsymbol{\theta}_j||_{\boldsymbol{\Lambda}_{i,t}}\\ 
    &+||\mathbf{A}_{i,t}^{-1}\boldsymbol{\xi}_{i,t}-\alpha \mathbf{A}_{i,t}^{-1}\sum_{j=1}^n \mathcal{L}_{ij}\mathbf{A}_{j,t}^{-1}\boldsymbol{\xi}_{j,t}||_{\boldsymbol{\Lambda}_{i,t}}\\
    \end{split}
\end{equation}
\begin{equation}
    \begin{split}
    &=\alpha ||\sum_{j=1}^n \mathcal{L}_{ij}\boldsymbol{\theta}_j||_{\mathbf{A}_{i,t}^{-1}\boldsymbol{\Lambda}_{i,t}\mathbf{A}_{i,t}^{-1}}\\
    &+||\boldsymbol{\xi}_{i,t}-\alpha \sum_{j=1}^n\mathcal{L}_{ij}\mathbf{A}_{j,t}^{-1}\boldsymbol{\xi}_{i,t}||_{\mathbf{A}_{i,t}^{-1}\boldsymbol{\Lambda}_{i,t}\mathbf{A}_{i,t}^{-1}}
    \end{split}
\end{equation}
Here we apply $||\mathbf{A}_{i,t}^{-1}(\cdot)||_{\boldsymbol{\Lambda}_{i,t}}=||\cdot||_{\mathbf{A}_{i,t}^{-1}\boldsymbol{\Lambda}_{i,t}\mathbf{A}_{i,t}^{-1}}$.
\begin{equation}
    \leq \alpha ||\sum_{j=1}^n \mathcal{L}_{ij}\boldsymbol{\theta}_j||_{\mathbf{A}_{i,t}^{-1}}
    +||\boldsymbol{\xi}_{i,t}-\alpha \sum_{j=1}^n\mathcal{L}_{ij}\mathbf{A}_{j,t}^{-1}\boldsymbol{\xi}_{i,t}||_{\mathbf{A}_{i,t}^{-1}}
\end{equation}
Here we use $||\cdot||_{\mathbf{A}_{i,t}^{-1}\boldsymbol{\Lambda}_{i,t}\mathbf{A}^{-1}_{i,t}}\leq ||\cdot||_{\mathbf{A}_{i,t}^{-1}}$.
\begin{equation}
\label{eq: noise_1}
    \leq \alpha ||\sum_{j=1}^n \mathcal{L}_{ij}\boldsymbol{\theta}_j||_{\mathbf{V}_{i,t}^{-1}}
    +||\boldsymbol{\xi}_{i,t}-\alpha \sum_{j=1}^n\mathcal{L}_{ij}\mathbf{A}_{j,t}^{-1}\boldsymbol{\xi}_{i,t}||_{\mathbf{V}_{i,t}^{-1}}
\end{equation}
Here we use $||\cdot||_{\mathbf{A}_{i,t}^{-1}}\leq ||\cdot||_{\mathbf{V}_{i,t}^{-1}}$.
\begin{equation}
\label{eq: noise_2}
    \leq \alpha ||\sum_{j=1}^n \mathcal{L}_{ij}\boldsymbol{\theta}_j||_{\mathbf{V}_{i,t}^{-1}}
    +||\boldsymbol{\xi}_{i,t}||_{\mathbf{V}_{i,t}^{-1}}
\end{equation}
Here we use 
\begin{equation}
\label{eq: noise_bound}
||\boldsymbol{\xi}_{i,t}-\alpha \sum_{j=1}^n\mathcal{L}_{ij}\mathbf{A}_{j,t}^{-1}\boldsymbol{\xi}_{i,t}||_{\mathbf{V}_{i,t}^{-1}} \leq ||\boldsymbol{\xi}_{i,t}||_{\mathbf{V}_{i,t}^{-1}}
\end{equation}
To support Eq.~\ref{eq: noise_bound}, we provide an empirical evidence in Figure~\ref{fig: noise_term_approximation}.

Denote $\boldsymbol{\Delta}_i=\sum_{j=1}^n\mathcal{L}_{ij}\boldsymbol{\theta}_j$, Finally, we have
\begin{equation}
    ||\hat{\boldsymbol{\theta}}_i-\boldsymbol{\theta}_i||_{\boldsymbol{\Lambda}_{i,t}}\leq \alpha ||\boldsymbol{\Delta}_i||_{\mathbf{V}_{i,t}^{-1}}
    +||\boldsymbol{\xi}_{i,t}||_{\mathbf{V}_{i,t}^{-1}}
\end{equation}

According to Theorem 2 in \cite{abbasi2011improved}, we have the following upper bound holds with probability $1-\delta$ for $\delta \in [0,1]$.
\begin{equation}
    ||\boldsymbol{\xi}_{i,t}||_{\mathbf{V}_{j,t}^{-1}}\leq \sigma \sqrt{2\log \frac{|\mathbf{V}_{i,t}|^{1/2}}{\delta|\alpha \mathbf{I}|^{1/2}}}
\end{equation}
In addition, we known
\begin{equation}
    \alpha ||\mathbf{\Delta}_i||_{\mathbf{V}_{i,t}^{-1}}\leq \sqrt{\alpha}||\mathbf{\Delta}_i||_2
\end{equation}
where we use $||\boldsymbol{\Delta}_i||^2_{\mathbf{V}_{i,t}^{-1}}\leq \frac{1}{\lambda_{min}}(\mathbf{V}_{i,t})||\boldsymbol{\Delta}_i||_2^2\leq \frac{1}{\alpha}||\boldsymbol{\Delta}_i||_2$, which means $\alpha||\boldsymbol{\Delta}_i||_{\mathbf{V}_{i,t}^{-1}}\leq \frac{\alpha}{\sqrt{\alpha}}||\boldsymbol{\Delta}_i||_2=\sqrt{\alpha}||\boldsymbol{\Delta}_i||_2$.

Finally, combine the above two upper bounds, we have the following upper bound holds with probability $1-\delta$ for $\delta \in [0,1]$.
\begin{equation}
\begin{split}
    ||\hat{\boldsymbol{\theta}}_{i,t}-\boldsymbol{\theta}_i||_{\boldsymbol{\Lambda}_{i,t}}
    &\leq \sigma \sqrt{2\log \frac{|\mathbf{V}_{i,t}|^{1/2}}{\delta|\alpha \mathbf{I}|^{1/2}}}+\sqrt{\alpha}||\boldsymbol{\Delta}_i||_2
\end{split}
\end{equation}

which means 
\begin{equation}
\label{eq: beta_proof}
\begin{split}
    \beta_{i,t}
    &=\sigma \sqrt{2\log \frac{|\mathbf{V}_{i,t}|^{1/2}}{\delta|\alpha \mathbf{I}|^{1/2}}}+\sqrt{\alpha}||\boldsymbol{\Delta}_i||_2
\end{split}
\end{equation}

\end{proof}

The green curve represents the LHS term in Eq.~\ref{eq: noise_bound}, while the blue curve represents the RHS term. Clearly, the blue curve is above the green curve and they converges together with large $t$.

\section*{Appendix E}
\textbf{Pseudocode of \textbf{GraphUCB-Local}}

\begin{algorithm}[ht]
\SetKwInOut{Input}{Input}\SetKwInOut{Output}{Output}
\Input{ $\alpha$, $T$, $\boldsymbol{\mathcal{L}}$, $\delta$}
\textbf{Initialization~~~:} For any $i \in \{1,2,...,n\}$ $\hat{\boldsymbol{\theta}}_{0,i}=\mathbf{0}\in \mathbb{R}^d$,~$\boldsymbol{\Lambda}_{0,i}=\mathbf{0}\in \mathbb{R}^{d \times d}$, $\mathbf{A}_{0,i}=\mathbf{0} \in \mathbb{R}^{d \times d}$, $\beta_{i,t}=0$.
\BlankLine
\For{$t \in [1,T]$}{
User index $i$ is selected
\begin{enumerate}
    \item $\mathbf{A}_{i,t} \gets \mathbf{A}_{i,t-1}+\mathbf{x}_{ t-1}\mathbf{x}_{t-1}^T$.
    \item $\mathbf{A}_{j,t}\gets \mathbf{A}_{j,t-1}$, $\forall j\neq i$.
    \item Update $\boldsymbol{\Lambda}_{i,t}$ via Eq. \ref{eq: single_precision}.
    \item Select $\mathbf{x}_{t}$ via Eq. \ref{eq: arm_selection}\\
    where $\beta_{i,t}$ is defined in Eq. \ref{eq: beta}.
    \item Receive the payoff $y_{t}$.
    \item Update $\hat{\boldsymbol{\theta}}_{i,t}$  via Lemma. \ref{lemma: single_solution} if $i=i_t$.
    \item $\hat{\boldsymbol{\theta}}_{j,t}\gets \hat{\boldsymbol{\theta}}_{j,t-1}$ \ $\forall j\neq i_t$.
    \end{enumerate}
}
\caption{\textbf{GraphUCB-Local}}
\label{algorithm:g-ucb-local}
\end{algorithm}

\section*{Appendix F} 
\textbf{Proof of Lemma \ref{lemma: psi}}:
Recall $\Psi_{i,T}=\frac{\sum_{\tau \in \mathcal{T}_{i,T}}||\mathbf{x}_{\tau}||^2_{\boldsymbol{\Lambda}_{i,\tau}^{-1}}}{\sum_{\tau \in \mathcal{T}_{i,T}}||\mathbf{x}_\tau||^2_{\mathbf{V}_{i,\tau}^{-1}}}$, where $\mathcal{T}_{i,T}$ is the set of time user $i$ is served up to time $T$, $\mathbf{A}_{i,\tau}=\sum_{\ell \in \mathcal{T}_{i,\tau}}\boldsymbol{x}_\ell \boldsymbol{x}_\ell^T$, $\mathbf{V}_{i,\tau}=\mathbf{A}_{i,\tau}+\alpha \mathcal{L}_{ii}\mathbf{I}$ and $\boldsymbol{\Lambda}_{i,T}$ defined in Eq.~\ref{eq: single_precision}. Without loss of generality, assume $||\mathbf{x}_\tau||_2\leq 1$ for any $\tau \leq T$.

\begin{proof}
\begin{equation}
\begin{split}
    \sum_{\tau \in \mathcal{T}_{i,T}}||\mathbf{x}_\tau||^2_{\mathbf{V}_{i,\tau}^{-1}}\leq (1+\max_{\tau \in \mathcal{T}_{i,T}}||\mathbf{x}_\tau||_2)\log |\mathbf{V}_{i,\tau}|
\end{split}
\end{equation}
In the same fashion
\begin{equation}
\begin{split}
    \sum_{\tau \in \mathcal{T}_{i,T}}||\mathbf{x}_{\tau}||^2_{\boldsymbol{\Lambda}_{i,\tau}^{-1}}\leq (1+\max_{\tau \in \mathcal{T}_{i,T}}||\mathbf{x}_\tau||_2)\log |\boldsymbol{\Lambda}_{i,\tau}|
\end{split}
\end{equation}
Since we assume $||\mathbf{x}_\tau||_2\leq 1$ for any $\tau \leq T$ and
\begin{equation}
    \Psi_{i,T}=\frac{\sum_{\tau \in \mathcal{T}_{i,T}}||\mathbf{x}_{\tau}||^2_{\boldsymbol{\Lambda}_{i,\tau}^{-1}}}{\sum_{\tau \in \mathcal{T}_{i,T}}||\mathbf{x}_\tau||^2_{\mathbf{V}_{i,\tau}^{-1}}}
\end{equation}
Given 
\begin{equation}\footnote{For isolated node, we set $\mathcal{L}_{ii}=1$, $\mathcal{L}_{ij}=0, j \neq i$.}
    \begin{split}
        \mathbf{V}_{i,\tau}=\mathbf{A}_{i,\tau}+\alpha \mathbf{I}\\
        \boldsymbol{\Lambda}_{i,\tau}=\mathbf{A}_{i,\tau}+2\alpha\mathcal{L}_{ii}\mathbf{I}+\alpha^2\sum_{j=1}^n \mathcal{L}_{ij}^2\mathbf{A}_{j,\tau}^{-1}
    \end{split}
\end{equation}
Assume\footnote{This can be ensured trivially by adding $\lambda \mathbf{I}$ to $\mathbf{A}_{i,\tau}$ with a small $\lambda$.} $\mathbf{A}_{i,\tau}$ (and $\mathbf{A}_{j,\tau}$) are positive semi-definite for all $\tau$. 
we know that $\boldsymbol{\Lambda}_{i,\tau}> \mathbf{V}_{i,\tau}$, therefore
\begin{equation}
    \boldsymbol{\Lambda}_{i,\tau}^{-1}< \mathbf{V}_{i,\tau}^{-1}
\end{equation}
holds for any $\tau \leq T$, which means
\begin{equation}
    ||\mathbf{x}_\tau||_{\boldsymbol{\Lambda}_{i,\tau}^{-1}}< ||\mathbf{x}_\tau||_{\mathbf{V}_{i,\tau}^{-1}}
\end{equation}
holds for any $\tau \leq T$. \\
Thus, we have 
\begin{equation}
    \sum_{\tau \in \mathcal{T}_{i,T}}||\mathbf{x}_\tau||^2_{\boldsymbol{\Lambda}_{i,\tau}^{-1}}< \sum_{\tau \in \mathcal{T}_{i,T}}||\mathbf{x}_\tau||^2_{\mathbf{V}_{i,\tau}^{-1}}
\end{equation}
this means 
\begin{equation}
    \Psi_{i,T}=\frac{\sum_{\tau \in \mathcal{T}_{i,T}}||\mathbf{x}_{\tau}||^2_{\boldsymbol{\Lambda}_{i,\tau}^{-1}}}{\sum_{\tau \in \mathcal{T}_{i,T}}||\mathbf{x}_\tau||^2_{\mathbf{V}_{i,\tau}^{-1}}}< 1
\end{equation}


In addition, since $||\mathbf{x}_\tau||^2_{\boldsymbol{\Lambda}_{i,\tau}^{-1}}>0$ and $||\mathbf{x}_\tau||^2_{\mathbf{V}_{i,\tau}^{-1}}>0$, This must hold
\begin{equation}
    \Psi_{i,T}>0
\end{equation}
Combine all together, we have 
\begin{equation}
    \Psi_{i,T}\in (0, 1)
\end{equation}
Furthermore, as $\mathbf{A}_{j,\tau}^{-1}$ decreases over time $\tau$, $\boldsymbol{\Lambda}_{i,\tau}\to \mathbf{A}_{i,\tau}$. It means $\Psi_{i,T}\to 1$.

\end{proof}

\section*{Appendix G}
\textbf{Proof of Theorem \ref{theorem: regret_upper_bound}}
\begin{proof}
First, we show the instantaneous regret at time $t$ can be upper bounded by $2\beta_{i,t}||\mathbf{x}_{i,t}||_{\boldsymbol{\Lambda}_{i,t}^{-1}}$ where $\mathbf{x}_{t}$ is the arm selected by the learner at time $t$ for user $i$. $\mathbf{x}_{i,*}$ is the optimal arm for user $i$.
\begin{equation}
    \begin{split}
        r_{i,t}
        &=\mathbf{x}^T_{i,*}\boldsymbol{\theta}_i-\mathbf{x}_{t}\boldsymbol{\theta}_i\\
        &\leq \mathbf{x}_{t}^T\hat{\boldsymbol{\theta}}_{i,t}+\beta_{i,t}||\mathbf{x}_{t}||_{\boldsymbol{\Lambda}_{i,t}^{-1}}-\mathbf{x}_{t}\boldsymbol{\theta}_i\\
        &\leq \mathbf{x}_{t}^T\hat{\boldsymbol{\theta}}_{i,t}+\beta_{i,t}||\mathbf{x}_{t}||_{\boldsymbol{\Lambda}_{i,t}^{-1}}-\mathbf{x}_{t}^T\hat{\boldsymbol{\theta}}_{i,t}+\beta_{i,t}||\mathbf{x}_{t}||_{\boldsymbol{\Lambda}_{i,t}^{-1}}\\
        &=2\beta_{i,t}||\mathbf{x}_{t}||_{\boldsymbol{\Lambda}_{i,t}^{-1}}
    \end{split}
\end{equation}
where we use the principle of optimistic
\begin{equation}
    \mathbf{x}_{i,*}^T\boldsymbol{\theta}_i+\beta_{i,t}||\mathbf{x}_{i,*}||_{\boldsymbol{\Lambda}_{i,t}^{-1}}\leq \mathbf{x}_{t}^T\hat{\boldsymbol{\theta}}_{i,t}+\beta_{i,t}||\mathbf{x}_{t}||_{\boldsymbol{\Lambda}_{i,t}^{-1}}
\end{equation}
and 
\begin{equation}
    \mathbf{x}_{t}^T\hat{\boldsymbol{\theta}}_{i,t}\leq \mathbf{x}_{t}^T\boldsymbol{\theta}_i+\beta_{i,t}||\mathbf{x}_{t}||_{\boldsymbol{\Lambda}_{i,t}^{-1}}
\end{equation}
Next, we drive a upper bound of the cumulative regret of user $i$ up to $T$
\begin{equation}
\label{eq: single_user_regret_bound}
    \begin{split}
        R_{i,T}=\sum_{\tau \in \mathcal{T}_{i,T}}r_{i,\tau}
        &\leq \sqrt{|\mathcal{T}_{i,T}|\sum_{\tau \in \mathcal{T}_{i,T}}r_{i,\tau}^2}\\
        &\leq \sqrt{|\mathcal{T}_{i,T}|\sum_{\tau \in \mathcal{T}_{i,T}}4\beta_{i,\tau}^2||\mathbf{x}_{\tau}||^2_{\boldsymbol{\Lambda}_{i,\tau}^{-1}}}\\
        &\leq 2\beta_{i,T}\sqrt{|\mathcal{T}_{i,T}| \sum_{\tau \in \mathcal{T}_{i,T}}||\mathbf{x}_\tau||^2_{\boldsymbol{\Lambda}_{i,\tau}^{-1}}}\\
        &\leq 2\beta_{i,T}\sqrt{|\mathcal{T}_{i,T}| \sum_{\tau \in \mathcal{T}_{i,T}}min(1, ||\mathbf{x}_\tau||^2_{\boldsymbol{\Lambda}_{i,\tau}^{-1}})}
    \end{split}
\end{equation}
where we user $\beta_{i,T}\geq \beta_{i,\tau}$ since $\beta_{i,\tau}$ is an increasing function over $\tau$ and $r_{i,\tau}\leq 2$ since we assume payoff $\mathbf{x}^T\boldsymbol{\theta}_i\in [-1,1]$.

According to Lemma 11 in \cite{abbasi2011improved}, we have 
\begin{equation}
\begin{split}
    \sum_{\tau \in \mathcal{T}_{i,T}}min(1, ||\mathbf{x}_\tau||^2_{\mathbf{V}_{i,\tau}^{-1}})
    &\leq 2 \log \frac{|\mathbf{V}_{i,T}|}{|\alpha \mathbf{I}|}\\
    & \leq 2\sqrt{d |\mathcal{T}_{i,T}| \log(\alpha+|\mathcal{T}_{i,T}|/d)}
\end{split}
\end{equation}
where $\mathbf{A}_{i,T}=\sum_{\tau \in \mathcal{T}_{i,T}}\mathbf{x}_{\tau}\mathbf{x}_{\tau}^T$, $\mathbf{V}_{i,T}=\mathbf{A}_{i,T}+\alpha\mathcal{L}_{ii} \mathbf{I}$ and $\mathcal{L}_{ii}=1$.

Recall in Lemma \ref{lemma: psi}, we define 
\begin{equation}
    \Psi_{i,T}=\frac{\sum_{\tau \in \mathcal{T}_{i,T}}||\mathbf{x}_\tau||^2_{\boldsymbol{\Lambda}_{i,\tau}^{-1}}}{\sum_{\tau \in \mathcal{T}_{i,T}}||\mathbf{x}_\tau||^2_{\mathbf{V}_{i,\tau}^{-1}}}
\end{equation}
Therefore
\begin{equation}
\label{eq: psi_term}
\begin{split}
   \sum_{\tau \in \mathcal{T}_{i,T}}min(1, ||\mathbf{x}_\tau||^2_{\boldsymbol{\Lambda}_{i,\tau}^{-1}})
   &= \Psi_{i,T}\sum_{\tau \in \mathcal{T}_{i,T}}\min (1, ||\mathbf{x}_\tau||^2_{\boldsymbol{\Lambda}_{i,\tau}^{-1}})\\
   &\leq 2\Psi_{i,T}\sqrt{d |\mathcal{T}_{i,T}|\log(\alpha+|\mathcal{T}_{i,t}|/d)} 
\end{split}
\end{equation}
From Eq.~\ref{eq: beta_proof}, we have
\begin{equation}
\begin{split}
    \beta_{i,T}
    &=\sigma \sqrt{2\log \frac{|\mathbf{V}_{i,T}|^{1/2}}{\delta|\alpha \mathbf{I}|^{1/2}}}+\sqrt{\alpha}||\boldsymbol{\Delta}_i||_2
\end{split}
\end{equation}
According to Theorem 2 in \cite{abbasi2011improved}, 
\begin{equation}
\begin{split}
    \sigma \sqrt{2\log \frac{|\mathbf{V}_{i,T}|^{1/2}}{\delta|\alpha \mathbf{I}|^{1/2}}}
    &\leq \sigma \sqrt{d\log \frac{1+|\mathcal{T}_{i,T}|/\alpha}{\delta}}\\
    &\leq \mathcal{O}(\sqrt{d\log{|\mathcal{T}_{i,T}|}})
\end{split}
\end{equation}
Hence, 
\begin{equation}
\label{eq: beta_bound_proof}
    \beta_{i,t}
    \leq \mathcal{O}(\sqrt{d\log{\mathcal{T}_{i,T}}}+\sqrt{\alpha}||\boldsymbol{\Delta}_i||_2)
\end{equation}
Combine this with Eq.~\ref{eq: psi_term} and Eq.~\ref{eq: single_user_regret_bound}, we have
\begin{equation}
\begin{split}
    R_{i,T}
    &\leq \mathcal{O}\bigg(\big(\sqrt{d\log{|\mathcal{T}_{i,T}|}}+\sqrt{\alpha}||\boldsymbol{\Delta}_i||_2\big)\times \\
    &\Psi_{i,T}\sqrt{d|\mathcal{T}_{i,T}|\log(|\mathcal{T}_{i,T}|)}\bigg)\\
    &\leq \Tilde{\mathcal{O}}(\Psi_{i,T}d\sqrt{|\mathcal{T}_{i,T}|})
\end{split}
\end{equation}
where the constant term $\sqrt{\alpha}||\boldsymbol{\Delta}_i||_2$ and logarithmic terms are hidden. 

Assume users are served uniformly, i.e., $\mathcal{T}_{i,T}=T/n$. Then, over the time horizon $T$, the total cumulative regret experienced by all users satisfies the following upper bound with probability $1-\delta$ with $\delta\in [0,1]$.
\begin{equation}
\begin{split}
  R_T
  &=\sum_{i=1}^nR_{i,T}=\sum_{i=1}^n\Tilde{\mathcal{O}}\bigg(\Psi_{i,T}d\sqrt{|\mathcal{T}_{i,T}|}\bigg)\\
  &=\sum_{i=1}^n\Tilde{\mathcal{O}}\bigg(\Psi_{i,T}d\sqrt{T/n}\bigg)\\
  &\leq\Tilde{\mathcal{O}}\bigg(nd\sqrt{T/n}\max_{i\in \mathcal{U}}\Psi_{i,T}\bigg)\\
  &=\Tilde{\mathcal{O}}\bigg( d\sqrt{Tn}\max_{i\in \mathcal{U}}\Psi_{i,T}\bigg) 
\end{split}
\end{equation}
\end{proof}

\section*{Appendix H}
Given $\boldsymbol{\mathcal{L}}=\mathbf{D}^{-1}\boldsymbol{\mathbf{L}}$ is the random-walk graph Laplacian, where $\boldsymbol{\mathbf{L}}=\mathbf{D}-\mathbf{W}$ is the combinatorial Laplacian. Recall 
$\mathbf{\Theta}=[\boldsymbol{\theta}_1, \boldsymbol{\theta}_2, ...,\boldsymbol{\theta}_n] \in \mathbb{R}^{n \times d}$ contains user features $\boldsymbol{\theta}_i \in \mathbb{R}^d$ in rows. 
Then, the quadratic Laplacian form can be expressed in the following way: 
\begin{equation}
    tr(\boldsymbol{\Theta}^T\boldsymbol{\mathcal{L}}\boldsymbol{\Theta})=\sum_{k=1}^d \sum_{i\sim j}\frac{1}{4}\big(\frac{W_{ij}}{D_{ii}}+\frac{W_{ji}}{D_{jj}}\big)\big(\Theta_{ik}-\Theta_{jk}\big)^2
\end{equation}

\begin{proof}
\begin{equation}
   tr(\boldsymbol{\Theta}^T\boldsymbol{\mathcal{L}}\boldsymbol{\Theta})=\sum_{k=1}^d \boldsymbol{\Theta}_{::k}^T\boldsymbol{\mathcal{L}}\boldsymbol{\Theta}_{::k}
\end{equation}
where $\boldsymbol{\Theta}_{::k} \in \mathbb{R}^n$ is the $k$-th column of $\boldsymbol{\Theta}$.

Note that $\boldsymbol{\mathcal{L}}=\mathbf{D}^{-1}\mathbf{L}$ is an asymmetric matrix with off-diagonal element $\mathcal{L}_{ij}=-\frac{W_{ij}}{D_{ii}}$ and $\mathcal{L}_{ji}=-\frac{W_{ji}}{D_{jj}}$ and on-diagonal element $\mathcal{L}_{ii}=1$.

From elementary linear algebra, we know that 
\begin{equation}
    \boldsymbol{\mathcal{L}}=\frac{\boldsymbol{\mathcal{L}}+\boldsymbol{\mathcal{L}}^T}{2}+\frac{\boldsymbol{\mathcal{L}}-\boldsymbol{\mathcal{L}}^T}{2}
\end{equation}
Then
\begin{equation}
\begin{split}
  tr(\mathbf{\Theta}^T\boldsymbol{\mathcal{L}}\mathbf{\Theta})
  &=tr(\mathbf{\Theta}^T(\frac{\boldsymbol{\mathcal{L}}+\boldsymbol{\mathcal{L}}^T}{2}+\frac{\boldsymbol{\mathcal{L}}-\boldsymbol{\mathcal{L}}^T}{2})\mathbf{\Theta})\\
  &=tr(\mathbf{\Theta}^T(\frac{\boldsymbol{\mathcal{L}}+\boldsymbol{\mathcal{L}}^T}{2})\mathbf{\Theta})+tr(\mathbf{\Theta}^T(\frac{\boldsymbol{\mathcal{L}}-\boldsymbol{\mathcal{L}}^T}{2})\mathbf{\Theta})\\
  &=tr(\mathbf{\Theta}^T(\frac{\boldsymbol{\mathcal{L}}+\boldsymbol{\mathcal{L}}^T}{2})\mathbf{\Theta})
\end{split}
\end{equation}
where 
\begin{equation}
tr(\mathbf{\Theta}^T(\frac{\boldsymbol{\mathcal{L}}-\boldsymbol{\mathcal{L}}^T}{2})\mathbf{\Theta})=0
\end{equation}
To see this
\begin{equation}
tr(\mathbf{\Theta}^T(\frac{\boldsymbol{\mathcal{L}}-\boldsymbol{\mathcal{L}}^T}{2})\mathbf{\Theta})=\sum_{k=1}^d \mathbf{\Theta}^T_{::k}(\frac{\boldsymbol{\mathcal{L}}-\boldsymbol{\mathcal{L}}^T}{2})\mathbf{\Theta}_{::k}
\end{equation}
for any $k \in \{1,2,..,d\}$
\begin{equation}
    \mathbf{\Theta}^T_{::k}(\frac{\boldsymbol{\mathcal{L}}-\boldsymbol{\mathcal{L}}^T}{2})\mathbf{\Theta}_{::k}=0
\end{equation}
To see this, 
assume $p=\mathbf{\Theta}^T_{::k}(\frac{\boldsymbol{\mathcal{L}}-\boldsymbol{\mathcal{L}}^T}{2})\mathbf{\Theta}_{::k}$, then
\begin{equation}
\begin{split}
    p
    &=\mathbf{\Theta}^T_{::k}(\frac{\boldsymbol{\mathcal{L}}-\boldsymbol{\mathcal{L}}^T}{2})\mathbf{\Theta}_{::k}\\
    &=\bigg(\mathbf{\Theta}^T_{::k}(\frac{\boldsymbol{\mathcal{L}}-\boldsymbol{\mathcal{L}}^T}{2})\mathbf{\Theta}_{::k}\bigg)^T\\
    &=-p
\end{split}
\end{equation}
where $\bigg(\frac{\boldsymbol{\mathcal{L}-\boldsymbol{\mathcal{L}^T}}}{2}\bigg)^T =-\frac{\boldsymbol{\mathcal{L}^T-\boldsymbol{\mathcal{L}}}}{2}$. 
So $p=0$.

Then,
\begin{equation}
    tr(\boldsymbol{\Theta}^T\boldsymbol{\mathcal{L}}\boldsymbol{\Theta})=tr(\boldsymbol{\Theta}^T(\frac{\boldsymbol{\mathcal{L}+\boldsymbol{\mathcal{L}}^T}}{2})\boldsymbol{\Theta})
\end{equation}
where the off-diagonal element, $i\neq j$ , of $\frac{\boldsymbol{\mathcal{L}}+\boldsymbol{\mathcal{L}}^T}{2}$ is $\frac{1}{2}\big(\frac{W_{ij}}{D_{ii}}+\frac{W_{ji}}{D_{jj}}\big)$ and the on-diagonal element is $1$.

Therefore
\begin{equation}
    tr(\boldsymbol{\Theta}^T\boldsymbol{\mathcal{L}}\boldsymbol{\Theta})=\sum_{k=1}^d \sum_{i\sim j}\frac{1}{4}\big(\frac{W_{ij}}{D_{ii}}+\frac{W_{ji}}{D_{jj}}\big)\big(\Theta_{ik}-\Theta_{jk}\big)^2
\end{equation}
\end{proof}

\section*{Appendix J}
To simulate $\mathcal{G}$, we follow two random graph models commonly used in the network science community: 1) Radial basis function (\textit{RBF}) model, a weighted fully connected graph, with edge weights $W_{ij}=exp(-\rho||\boldsymbol{\theta_i}-\boldsymbol{\theta_j}||^2)$; 2) Erd\H{o}s R\'enyi (ER) model, an unweighted graph, in which each edge is generated independently and randomly with probability $p$. 
3) \textit{Barab\'asi-Albert} (BA) model, an unweighted graph
initialized with a connected graph with $m$ nodes. Then, a new node is added to the graph sequentially with $m$ edges connected to existing nodes following the rule of preferential attachment where existing nodes with more edges has more probability to be connected by the new node; 4) \textit{Watts-Strogatz} (WS) model, an unweighted graph, which is a $m$-regular graph with edges randomly rewired with probability $p$. 
For each graph model, different topologies can be generated, leading to different level of sparsity and smoothness as show in the following 

\begin{figure}[ht]
    \centering
    \subfigure[][RBF (s=0.5)]{
        \includegraphics[width=1.5in]{figures/rbf_cum_regret.png}
        }
    \subfigure[][ER (p=0.4)]{
        \includegraphics[width=1.5in]{figures/er_cum_regret.png}
        }
    \subfigure[][BA (m=5)]{
        \includegraphics[width=1.5in]{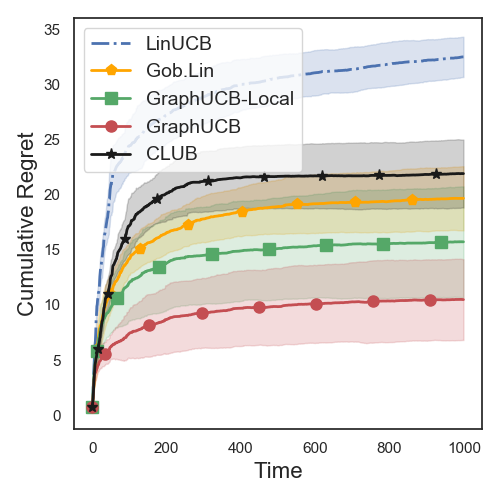}
        }
    \subfigure[][WS (m=7, p=0.2)]{
        \includegraphics[width=1.5in]{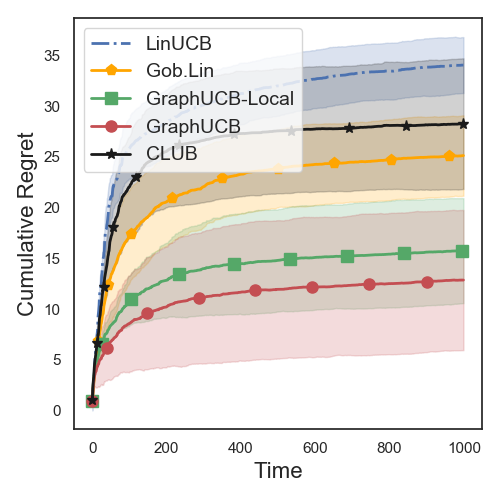}
        }
\caption{Performance with respect to Graph Types}
\label{fig: performance_on_graph_structures_weighted_appendix}
\end{figure}

\begin{figure}[ht]
    \centering
    \subfigure[][GraphUCB]{
        \includegraphics[width=1.5in]{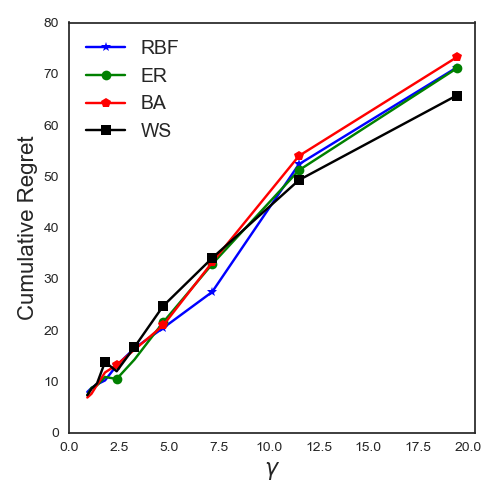}
        }
    \subfigure[][GraphUCB-Local]{
        \includegraphics[width=1.5in]{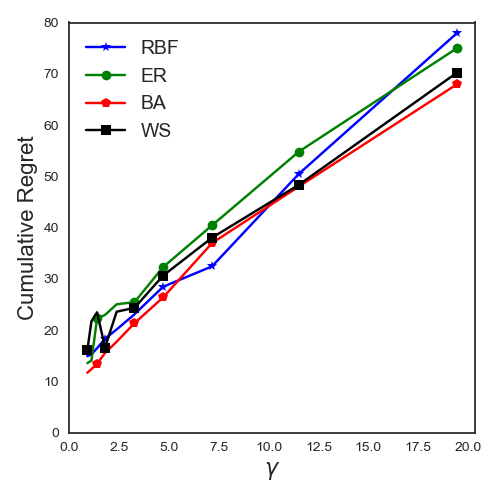}
        }
\caption{Performance of proposed algorithms on Graph models}
\label{fig: graphucb_smooth_graphs_weighted_appendix}
\end{figure}
\begin{figure}[ht]
    \centering
    \subfigure[][RBF (s=0.5)]{
        \includegraphics[width=1.5in]{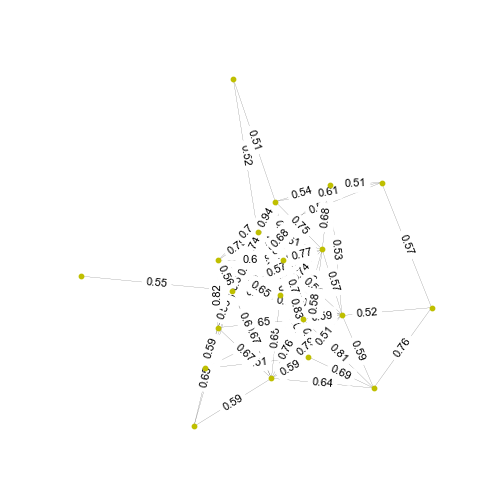}
        }
    \subfigure[][ER (p=0.4)]{
        \includegraphics[width=1.5in]{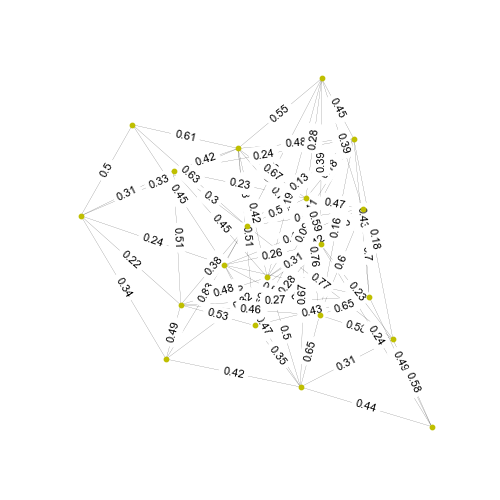}
        }
    \subfigure[][BA (m=5)]{
        \includegraphics[width=1.5in]{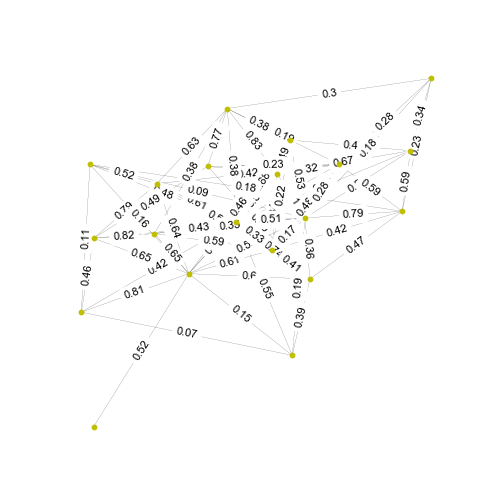}
        }
    \subfigure[][WS (m=7, p=0.2)]{
        \includegraphics[width=1.5in]{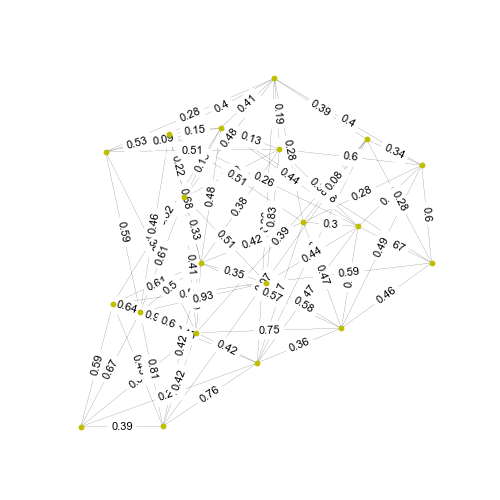}
        }
\caption{Graphs}
\label{fig: graph_structures}
\end{figure}

Comparing sub-figures in Figure \ref{fig: performance_on_graph_structures_weighted_appendix} shows that with the same level of sparsity and smoothness the effect of topology of graph performance seems to be unnoticeable. The generated graphs are shown in Figure~\ref{fig: graph_structures}. This is also confirmed in Figure~\ref{fig: graphucb_smooth_graphs_weighted_appendix}, in which we generate graphs with different topology but with the same level connectivity. Algorithms are test on smoothness level. 

\section*{Appendix L}
We test the performance of the proposed algorithms on the basis of graph properties such as smoothness, sparsity, $p$ in ER graph, $m$ in BA graph, $m$ and $p$ in WS graph.
\newline
\textbf{Smoothness [$\gamma$]:}
We first generate a \textit{RBF} graph. To control the smoothness, we vary $\gamma \in [0,10]$. The term $sm=tr(\mathbf{\Theta}^T\boldsymbol{\mathcal{L}\mathbf{\Theta}})$ measures the corresponding smoothness level. To ensure the comparison is fair, $\mathbf{\Theta}$ is normalized to be $||\mathbf{\Theta}||_2=n$.
\newline
\textbf{Sparsity [$s$]:}
We first generate a \textit{RBF} graph, then generate a smooth $\mathbf{\Theta}$ via Eq.~\ref{eq: smooth_signal}.
To control the sparsity, we set a threshold $s \in [0,1]$ on edge weights $W_{ij}$ such that $W_{ij}$ less than $s$ are removed. The term $sp=\frac{number of edges}{n(n-1)}$ measures the corresponding level of sparsity, where $n(n-1)$ is the number of edges of a fully connected graph.

\begin{figure}[ht]
    \centering
    \subfigure[][RBF ($\gamma$)]{
        \includegraphics[width=1.5in]{figures/smooth_rbf.png}
        }
    \subfigure[][RBF (s)]{
        \includegraphics[width=1.5in]{figures/threshold_rbf.png}
        }
    \subfigure[][ER (p)]{
        \includegraphics[width=1.5in]{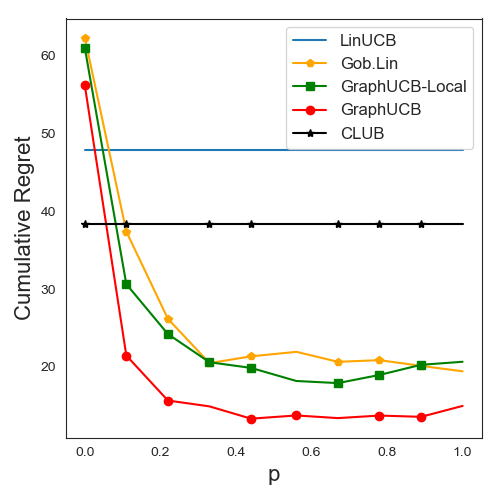}
        }
    \subfigure[][BA (m)]{
        \includegraphics[width=1.5in]{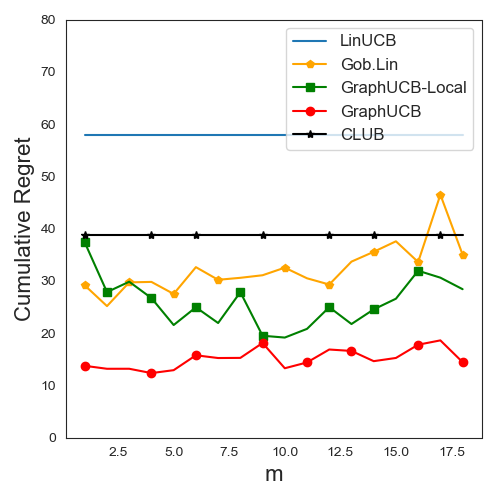}
        }
    \subfigure[][WS (p, m=4)]{
        \includegraphics[width=1.5in]{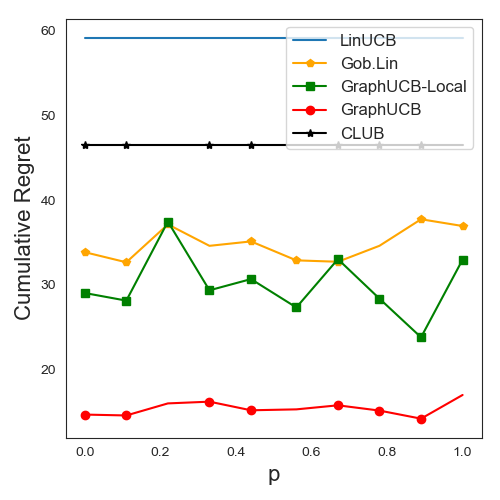}
        }
    \subfigure[][WS (m, p=0.2)]{
        \includegraphics[width=1.5in]{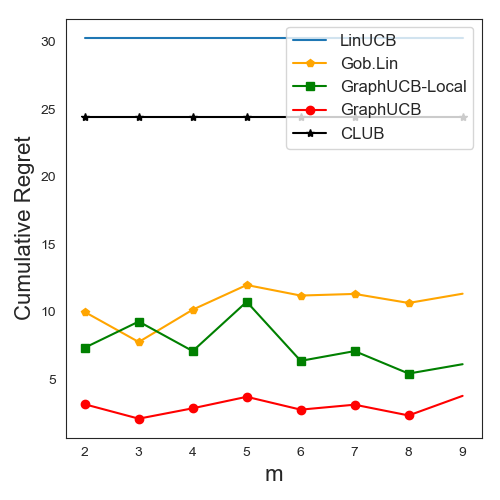}
        }
\caption{Performance on Graph Properties}
\label{fig: performance_on_graph_properties_appendix}
\end{figure}

Results are shown in Figure \ref{fig: performance_on_graph_properties_appendix}. In sub-figure (a) and (b), graph-based algorithms show similar pattern. Smoother signal leads to less regret because of the Laplacian-regularier estimator in Eq.~\ref{eq: objective_function}. Sparse graph leads to more regret as less connectivity provides less graph information. This is also confirmed by sub-figure (c), $p$ in ER controls the probability of edge. Small $p$ leads to spare graph, in turn more regret. 

\section*{Appendix K}
\textbf{Movilens} contains 6k users and their ratings on 40k movies. Since every user does not give ratings on all movies, there are a large mount of missing ratings.
 We factorize the rating matrix via $\mathbf{M}=\mathbf{U}\mathbf{X}$ to fill the missing values where $\mathbf{U}$ contain users' latent vectors in rows and $\mathbf{X}$ contain movies' latent features in columns. The dimension is set as $d=10$. Next, we create the user graph $\mathcal{G}$ from $\mathbf{U}$ via \textit{RBF} kernel. \textbf{Netflix} contains rating of 480k users on 18k movies. We process the dataset in the same way as \textbf{Movielens}. In both datasets, original ratings range from 0 to 5, we normalize them into $[0,1]$. After the data pre-processing, 
we sample 50 users and test algorithms over $T=1000$.
\begin{figure}[ht]
    \centering
    \subfigure[][MovieLens]{
        \includegraphics[width=1.5in]{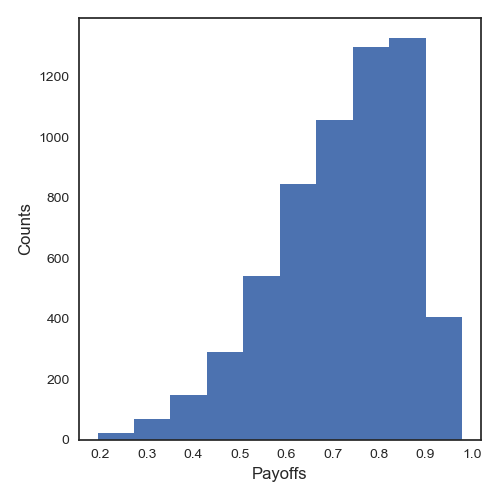}
        }
    \subfigure[][Netflix]{
        \includegraphics[width=1.5in]{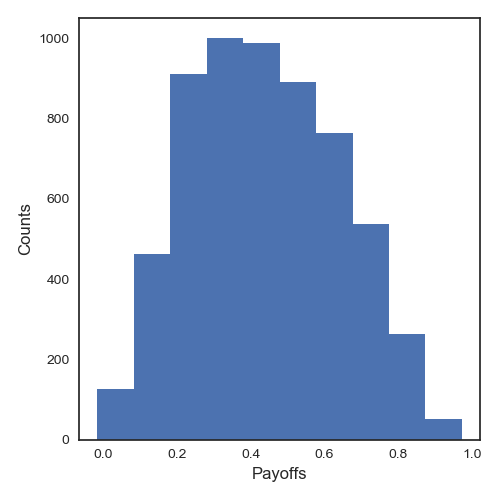}
        }
\caption{Histogram of signals in MovieLens (a) and Netflix (b).}
\label{fig: histogram_of_signals}
\end{figure}

Figure~\ref{fig: histogram_of_signals} shows the distribution of ratings in \textbf{Movielens} and \textbf{Netflix}. Ratings in \textbf{Movielens} are highly concentrated which means a large number of users like a few set of movies. They show similar performance. 
\end{document}